\documentclass[sigconf]{acmart}
\AtBeginDocument{%
  }

\copyrightyear{2026}
\acmYear{2026}
\setcopyright{cc}
\setcctype{by-nc-nd}
\acmConference[KDD '26]{Proceedings of the 32nd ACM SIGKDD Conference on Knowledge Discovery and Data Mining V.2}{August 09--13, 2026}{Jeju Island, Republic of Korea}
\acmBooktitle{Proceedings of the 32nd ACM SIGKDD Conference on Knowledge Discovery and Data Mining V.2 (KDD '26), August 09--13, 2026, Jeju Island, Republic of Korea}
\acmDOI{10.1145/3770855.3818173}
\acmISBN{979-8-4007-2259-2/2026/08}
\usepackage{hyperref}
\usepackage{url}

\usepackage{graphicx}
\usepackage[flushleft]{threeparttable}
\usepackage{multirow}
\usepackage{subcaption}
\usepackage{wrapfig}

\usepackage{float}
\usepackage{algorithm}
\usepackage{algorithmic}  
\usepackage{enumitem}

\usepackage{mathtools}
\usepackage{amsthm}
\usepackage{bm} 
\usepackage[capitalize,noabbrev]{cleveref}
\usepackage{booktabs}       % professional-quality tables
\usepackage{microtype}      % microtypography
\usepackage{amsfonts}       
\usepackage{amsmath} 
 
\usepackage{amssymb} 
\usepackage[table]{xcolor} 
\usepackage{balance}

\begin{document}
\title{Learning to Reduce Search Space for Generalizable \\ Neural Routing Solver}

\author{Changliang Zhou}
\authornote{Equal contribution}
\email{zhoucl2022@mail.sustech.edu.cn}
\orcid{0009-0005-4512-9558}
\affiliation{%
    \department{School of Automation and Intelligent Manufacturing}
    \institution{Southern University of Science and Technology}
    \city{Shenzhen}
    \state{Guangdong}
    \country{China}
}
\additionalaffiliation{
 \department{Guangdong Provincial Key Laboratory of Fully Actuated System Control Theory and Technology}
    \institution{Southern University of Science and Technology}
    \city{Shenzhen}
    \state{Guangdong}
    \country{China}
}

\author{Xi Lin}
\authornotemark[1]
\email{xi.lin@xjtu.edu.cn}
\orcid{0000-0001-5298-6893}
\affiliation{%
\department{School of Mathematics and Statistics}
  \institution{Xi'an Jiaotong University}
  \city{Xi'an}
  \state{Shaanxi}
  \country{China}
}

\author{Zhenkun Wang}
\email{wangzhenkun90@gmail.com}
\orcid{0000-0003-1152-6780}
\authornote{Corresponding author}
\affiliation{%
  \department{School of Automation and Intelligent Manufacturing}
  \institution{Southern University of Science and Technology}
  \city{Shenzhen}
  \state{Guangdong}
  \country{China}
}
\additionalaffiliation{
 \department{Guangdong Provincial Key Laboratory of Fully Actuated System Control Theory and Technology}
    \institution{Southern University of Science and Technology}
    \city{Shenzhen}
    \state{Guangdong}
    \country{China}
}

\author{Qingfu Zhang}
\email{qingfu.zhang@cityu.edu.hk}
\orcid{0000-0003-0786-0671}
\affiliation{%
\department{Department of Computer Science}
 \institution{City University of Hong Kong}
 \city{Hong Kong SAR}
 \country{China}
 }

\renewcommand{\shortauthors}{Zhou et al.}

\begin{abstract}
Constructive neural combinatorial optimization (NCO) offers a promising paradigm for solving vehicle routing problems (VRPs) by directly learning to construct approximate optimal solutions, thereby reducing reliance on expert knowledge for algorithm design. However, scaling these methods to handle large-scale instances remains challenging due to high computational complexity. While recent dynamic search space reduction (SSR) methods can improve inference efficiency through geometric distance-based pruning, they often struggle on complex instances with non-uniform distributions or when optimal solutions rely heavily on non-spatial constraints. To address this critical issue, we propose Learning to Reduce (L2R), which is the first learning-based dynamic SSR framework. L2R learns to adaptively prioritize nodes by extracting patterns from problem-specific features to prune the search space at each step, enabling efficient and scalable solution construction. Extensive experiments show that our L2R framework generalizes robustly to different problem scales and data distributions on various VRP variants. To the best of our knowledge, L2R is the first neural solver to effectively scale to VRP instances with $10$ million nodes while maintaining high solution quality, which significantly pushes the frontier of NCO in terms of generalization and scalability. Our code is available at \url{https://github.com/CIAM-Group/L2R}.
\end{abstract}


% %% information and the body of the document, and typically spans the


\maketitle
\newcommand\kddavailabilityurl{https://doi.org/10.5281/zenodo.20493139}
\ifdefempty{\kddavailabilityurl}{}{
\begingroup\small\noindent\raggedright\textbf{Resource Availability:}\\
% please change the following context to include multiple artifacts if necessary, including data, models, code, etc.
The source code of this paper has been made publicly available at \url{\kddavailabilityurl}.
\endgroup
}

\section{Introduction}
\label{sec:intro}
The Vehicle Routing Problem (VRP) is a core problem in Operations Research with significant practical implications across domains such as logistics, supply chain management, and express delivery~\citep{tiwari2023optimization,sar2023systematic}. Efficient routing optimization is crucial for improving delivery performance and reducing operational costs. Traditional heuristic algorithms, such as LKH3~\citep{LKH3} and HGS~\citep{HGS}, have demonstrated strong capabilities in solving VRPs with diverse constraints. However, these methods face two fundamental limitations: (1) their design requires extensive domain expertise to develop problem-specific rules, and (2) their computational complexity grows prohibitively with instance size due to the NP-hard nature of VRPs. These challenges are particularly critical for large-scale instances (e.g., with more than $10,000$ nodes), where existing algorithms often fail to produce practical solutions with reasonable runtime.

In recent years, neural combinatorial optimization (NCO) methods have attracted substantial attention for their potential to reduce reliance on handcrafted rules while maintaining competitive performance for solving VRPs~\citep{bengio2021machine,li2022overview,wu2024neural_survey,ba2026survey}. These methods learn problem-specific patterns automatically through training frameworks such as supervised learning (SL)~\citep{vinyals2015pointer,luo2023lehd,drakulic2023bq,xiao2023distilling,joshi2019efficient,hudson2021graph_gls,luo2026learning,luo2026efficient} or reinforcement learning (RL)~\citep{bello2016neural,khalil2017s2v_dqn,kool2019attention,zhou2024icam,zhou2026urs}. A well-trained NCO model can directly construct approximate optimal solutions without explicit search, offering a promising direction for real-time VRP solving. However, SL-based methods face a critical difficulty of obtaining high-quality labeled data (e.g., near-optimal solutions) for large-scale NP-hard problems. In contrast, RL-based methods do not require labeled data and have demonstrated strong performance on small-scale instances (e.g., with 100 nodes)~\citep{kwon2020pomo,kim2022symnco,xiao2025improving}. Nevertheless, their effectiveness diminishes significantly on large-scale instances, primarily due to the exponentially growing search space and the challenge of sparse rewards. 
\begin{table}[t]
    \centering
    \caption{Comparison of our L2R and classical neural routing solvers with search space reduction.}
    \resizebox{0.92\linewidth }{!}{
    \begin{threeparttable}
    \begin{tabular}{c|c|c|c|c}
\toprule[0.5mm]
Neural Routing  & Static   &  Dynamic &Training & Generalizable      \\ 
Solver & SSR & SSR  &Scale & Scale\\ 
\midrule
MLPR~\citep{sun2021MLPR}& $\checkmark$ & $\times$ &$100$ & 2K\\
Att-GCN+MCTS~\citep{fu2021attgcn_mcts}& $\checkmark$ & $\times$ &$50$ & 10K\\
DIMES~\citep{qiu2022dimes} & $\checkmark$ & $\times$ &10K  & 10K \\
DIFUSCO~\citep{sun2023difusco} & $\checkmark$ & $\times$& 10K  & 10K \\
T2T~\citep{li2024T2T} & $\checkmark$ & $\times$&  1K & 1K \\
BQ~\citep{drakulic2023bq}$\ddagger$ &$\times$ &Distance-based &$100$ & 1K \\
ELG~\citep{gao2023elg} &$\times$ &Distance-based &$100$ & 7K\\
DAR~\citep{wang2024distance} &$\times$ &Distance-based &$500$ & 11K\\
INViT~\citep{fang2024invit} &$\times$ &Distance-based &$100$ & 10K\\
\midrule
L2R (Ours) &\cellcolor[HTML]{D0CECE}\textbf{$\checkmark$} & \cellcolor[HTML]{D0CECE}\textbf{Learning-based} & \cellcolor[HTML]{D0CECE}\textbf{100} & \cellcolor[HTML]{D0CECE}\textbf{10M} \\
\toprule[0.5mm]
\end{tabular}
\begin{tablenotes}
\item[ $\ddagger$] BQ~\citep{drakulic2023bq} limits the sub-graph to the $250$ nearest neighbors of the current node when facing large-scale instances.
\end{tablenotes}
\end{threeparttable}
}
\vspace{-15pt}
\label{table:motivation}
\end{table}
To tackle the scalability challenge, search space reduction (SSR) has been adopted as a key strategy. As listed in Table \ref{table:motivation}, existing SSR techniques can be broadly categorized into two types: static and dynamic. Static SSR performs a one-time pruning step at the beginning of the process, thereby improving computational efficiency. However, it often requires additional search procedures to achieve high-quality solutions~\citep{fu2021attgcn_mcts,qiu2022dimes,sun2023difusco}. In contrast, dynamic SSR~\citep{fang2024invit,gao2023elg,wang2024distance} adaptively updates the candidate node set at each step based on real-time problem states, enabling more effective reduction for constructive methods. However, a key limitation of current dynamic SSR methods is their reliance on geometric distance. This makes them struggle to generalize to large-scale instances with non-uniform distributions or non-spatial constraints. A comprehensive literature review is available in Appendix \ref{append:related_work}. 

Unlike prior SSR methods that rely on geometric distance as a rigid hard-pruning rule, this work shifts towards a learned soft prioritization. Specifically, we propose \textit{Learning to Reduce (L2R)}, a novel dynamic SSR framework for large-scale VRP instances that adaptively integrates static distance priors with learned context-aware features. Our contributions are summarized as follows:  
\begin{itemize}
    \item We provide a comprehensive analysis of existing distance-based dynamic SSR methods and highlight their key limitations in solving large-scale VRP instances.
    \item We propose the first learning-based dynamic SSR framework that prunes the search space at each construction step by adaptively prioritizing nodes based on patterns learned from problem-specific features.
    \item Extensive experiments show that L2R achieves robust generalization across three VRPs while greatly reducing computational overhead without compromising solution quality. 
    \item To the best of our knowledge,  L2R is the first neural solver capable of handling VRP instances with up to $10$ million nodes while maintaining high-quality solutions, which significantly pushes the frontier of NCO in terms of generalization and scalability. 
\end{itemize}   

\section{Motivation and Key Idea}
\label{sec:preliminary}
The fundamental limitation of distance-based SSR stems from its tendency to prematurely prune globally optimal nodes during solution construction. In this section, we systematically analyze this behavior by examining its impact across three critical aspects: (1) the optimality gap in classical solvers, (2) constructive NCO methods with diverse architectures, and (3) VRP variants with complex constraints. Then, we propose a learning-based reduction framework to address this fundamental shortcoming.
\begin{table}[t]
\centering
\caption{Optimality gap comparison of different TSPLib instances using LKH-3 on unreduced and reduced search space. }
\resizebox{0.92\linewidth}{!}{
\begin{tabular}{l  | c| c | c|c }
\toprule[0.5mm]
Instance& Scale &w/o D-SSR &w/ $k=10$  & w/ $k=20$ \\
\midrule
dsj1000    &1,000  & 0.00\% & 7.27\% & 6.85\% \\
pr1002    & 1,002 & 0.00\% & 3.14\% & 1.06\% \\
d1291    &1,291  & 0.00\% & 26.71\% & 9.59\% \\
fl1400        &1,400  & 0.02\% & 64.72\% & 54.29\% \\
fl1577        &1,577  & 0.01\% & 48.14\% & 38.19\% \\
 d1655       & 1,655  & 0.00\% & 20.41\% & 8.74\% \\
rl1889        &1,889  & 0.00\% & 27.94\% & 8.45\% \\
d2103        & 2,103 & 0.01\% & 17.62\% & 6.10\% \\
fl3795 & 3795 & 0.06\% & 56.87\% & 53.38\% \\
rl5915 &5,915 & 0.03\% & 17.99\% & 2.58\% \\
rl5934 &5,934 & 0.03\% & 27.24\% & 6.40\%  \\
rl11849 &11,849 &   0.00\% & 11.01\% & 1.90\% \\
usa13509 &13,509  &  0.01\% & 20.06\% & 5.87\% \\
\midrule
\multicolumn{2}{c|}{Avg. Gap}&  0.01\% & 26.86\% & 15.64\% \\
\bottomrule[0.5mm]
\end{tabular}
}
\vspace{-12pt}
\label{table:TSPLIB_Optimal_ratio}
\end{table}
\begin{figure}[t]
    \centering
    \begin{subfigure}{0.235\textwidth}
        \centering
        \includegraphics[width=\textwidth]{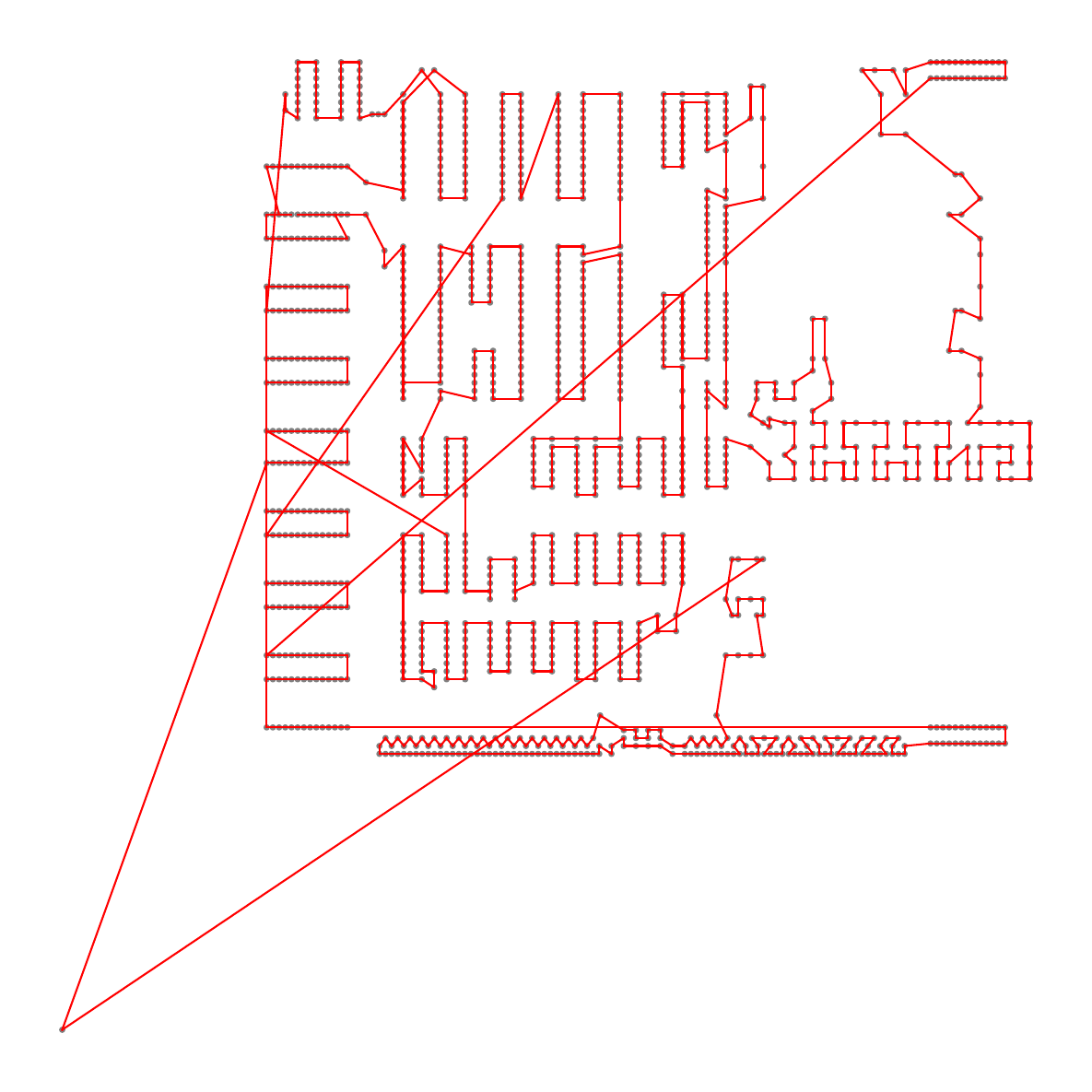}
        \caption{\centering $k=10$ (Gap:26.71\%)}
        \label{subfig:d1291_LKH_K10_solution}
    \end{subfigure}
    \hfill
     \begin{subfigure}{0.235\textwidth}
        \centering
        \includegraphics[width=\textwidth]{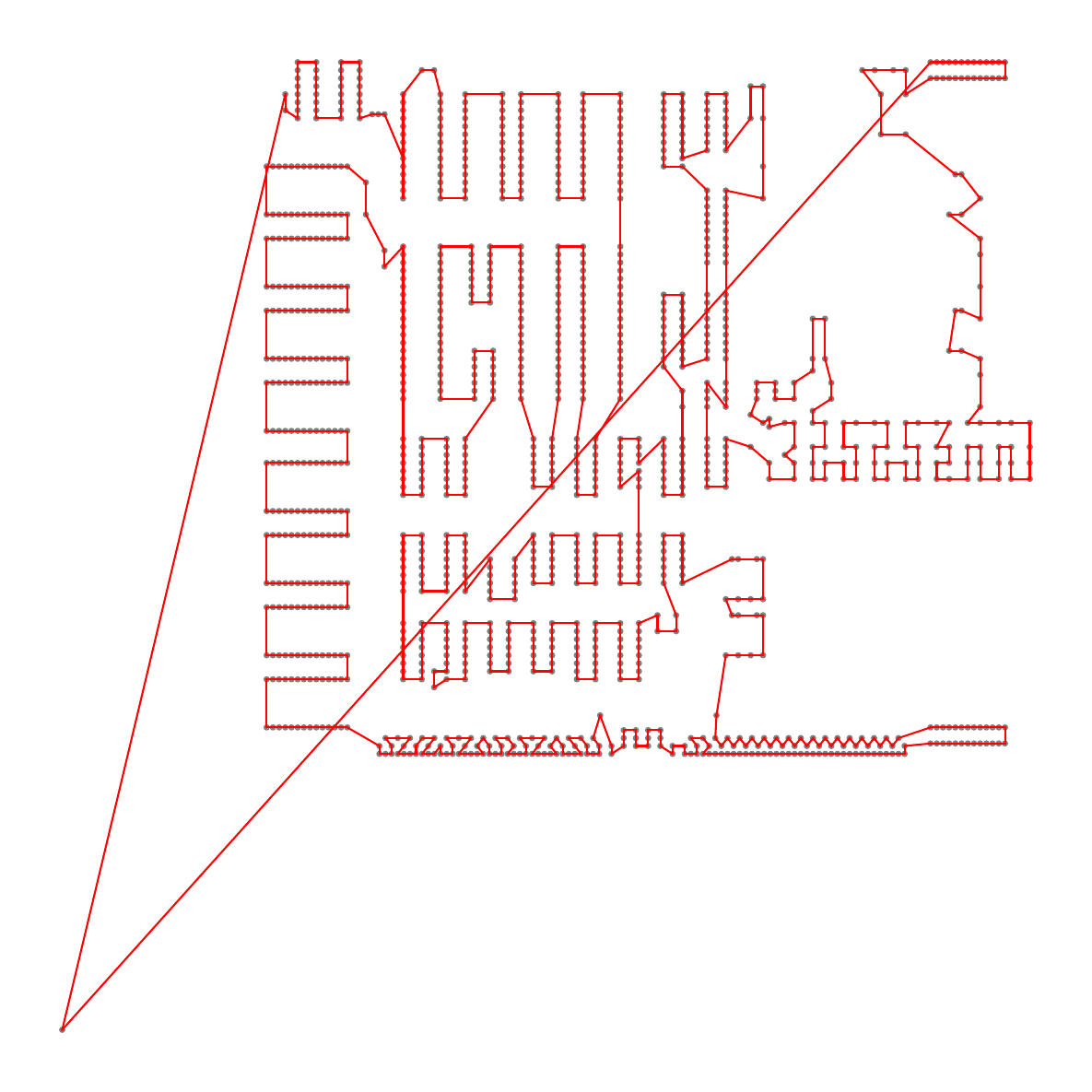}
        \caption{\centering $k=20$ (Gap:9.59\%)}
        \label{subfig:d1291_LKH_K20_solution}
    \end{subfigure}
\caption{Solution visualizations for instance d1291 using LKH-3~\citep {LKH3} under distance-based SSR with different $k$.}
\label{fig:tsplib_visualizations_lkh3_knn_mainpaper}
\vspace{-15pt}
\end{figure}

\subsection{Degradation of Solution Optimality}
To evaluate the impact of distance-based SSR on final optimization performance, we adopt the widely used LKH-3~\citep{LKH3} as our benchmark solver and conduct experiments on TSPLib~\citep{reinelt1991tsplib}. Given that LKH-3 is not a constructive heuristic, we restrict its search space for each instance by pruning the original fully connected graph into a sparse topology. Specifically, only connections to the $k$ nearest nodes are retained for each node in our experiments.

\cref{table:TSPLIB_Optimal_ratio} contains the results across TSPLib instances of varying scales under different levels of search space reduction ($k$). A key observation is that aggressive pruning (i.e., using small $k$ values) will substantially degrade performance, even though LKH-3 can obtain near-optimal solutions in the original search space. This degradation is particularly pronounced when optimal routes depend on non-local node selections, which are often eliminated by aggressive pruning strategies. As shown in Figure~\ref{fig:tsplib_visualizations_lkh3_knn_mainpaper}, restricting candidate nodes to the $k$-nearest neighbors forces the solver to ignore critical long-distance visits necessary for optimal routes. This over-pruning effect accumulates systematically throughout the solving process, ultimately compromising solution quality.
\begin{figure}[t]
    \centering
    \begin{subfigure}{0.47\textwidth}
        \centering
        \includegraphics[width=\textwidth]{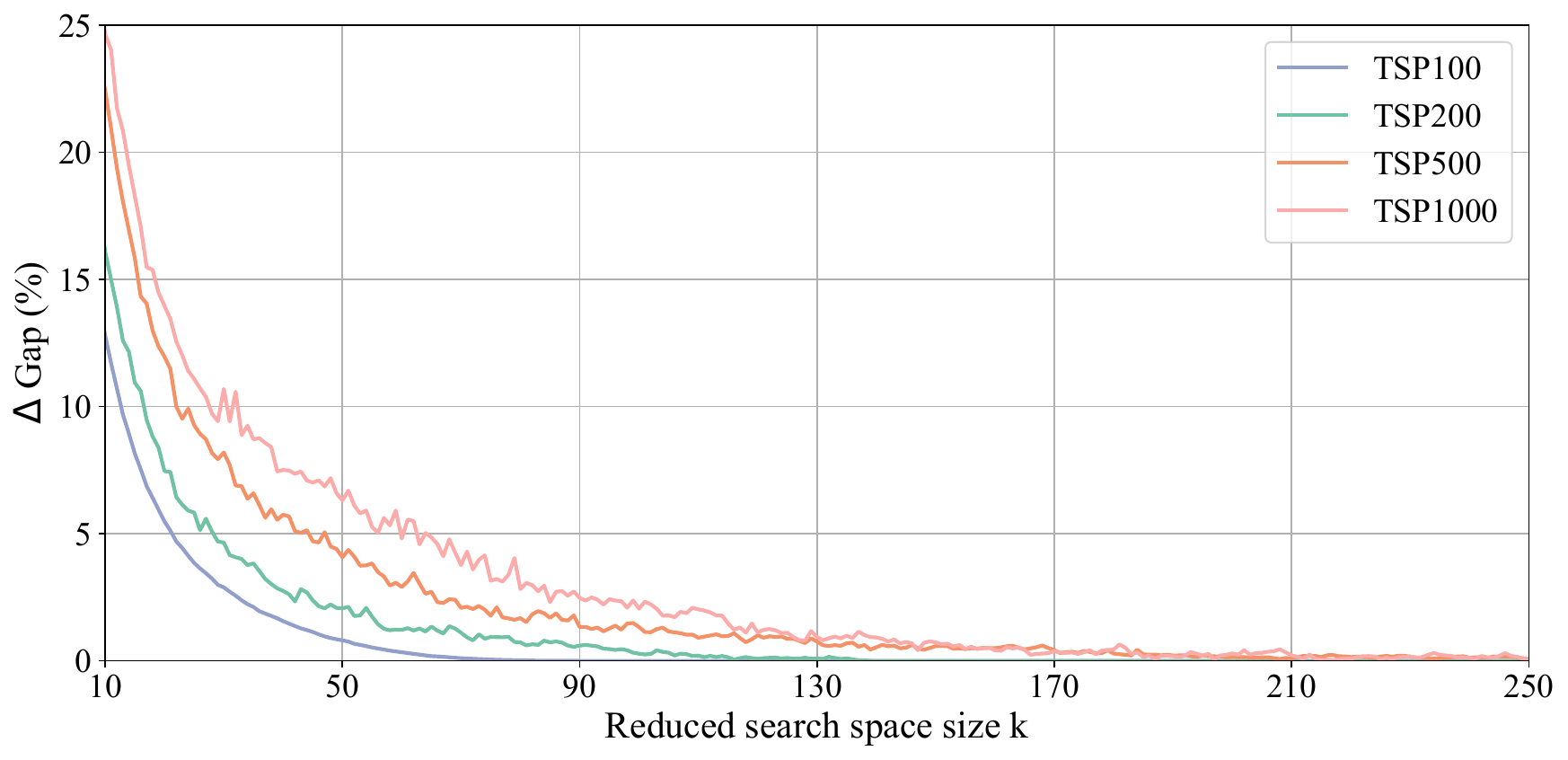}
        \caption{\centering LEHD with Light Encoder and Heavy Decoder.}
        \label{subfig:tsp_k_statistic_lehd}
    \end{subfigure}
    \hfill
     \begin{subfigure}{0.47\textwidth}
        \centering
        \includegraphics[width=\textwidth]{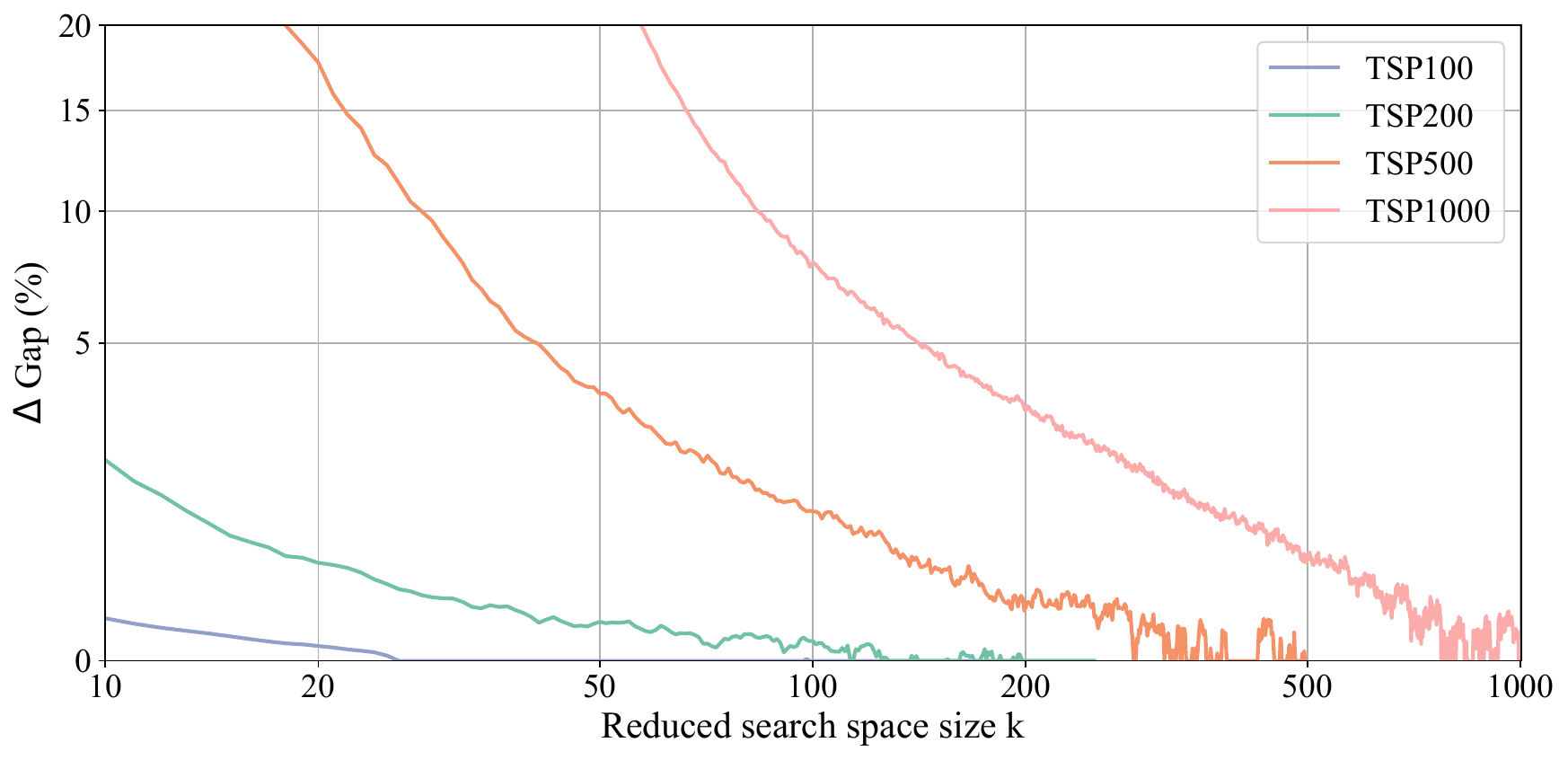}
        \caption{\centering POMO with Heavy Encoder and Light Decoder.}
        \label{subfig:tsp_k_statistic_pomo}
    \end{subfigure}
    \caption{Effects of distance-based SSR in diverse pretrained constructive models. Here $\Delta Gap=Gap_{\text{reduced}}-Gap_{\text{original}}$. }
    \label{fig:tsp_k_statistic_ncos}
    \vspace{-10pt}
\end{figure}
\subsection{Sensitivity of Constructive NCOs}
For constructive NCO methods, each step should consider all feasible nodes to preserve optimality. However, the search space of VRPs grows exponentially with problem size, making it extremely challenging to obtain high-quality solutions directly for large-scale instances. To evaluate the impact of distance-based SSR on diverse NCO architectures, we conduct experiments on three settings: (1) a pretrained constructive NCO with a heavy decoder, (2) a pretrained constructive NCO with a heavy encoder, and (3) a retrained constructive NCO.

\paragraph{Impact on Pretrained Constructive NCO with Heavy Decoder}
We use the well-known LEHD~\citep{luo2023lehd} as an example, restricting the search space to the $k$ nearest nodes to the last-visited node at each construction step. Without loss of generality, we use the LEHD without any RRC strategy in this experiment. We quantify the impact by measuring the performance gap $\Delta Gap=Gap_{\text{reduced}}-Gap_{\text{original}}$, defined as the difference between the optimality gap obtained by LEHD with a reduced search space ($Gap_{\text{reduced}}$) and the original LEHD without SSR operation ($Gap_{\text{original}}$).

The results in Figure~\ref{subfig:tsp_k_statistic_lehd} illustrate the performance on TSP instances of varying sizes under different search space reduction levels ($k$). A key observation is the existence of a critical threshold $k^*$ for each problem size, beyond which LEHD with $k \geq k^*$ achieves a $0\%$ performance gap with the original LEHD, indicating no loss in solution quality with SSR. However, these critical thresholds vary significantly across problem sizes, and using a small $k < k^*$ leads to substantial performance degradation. 

\paragraph{Impact on Pretrained Constructive NCO with Heavy Encoder}
Additionally, we evaluate the widely studied POMO model with a heavy encoder architecture and remove instance augmentation in our experiment. As shown in \cref{subfig:tsp_k_statistic_pomo}, our experimental results demonstrate a consistent phenomenon with the LEHD case: The critical thresholds $k^*$ exhibit significant variations across different problem sizes, and using a small $k < k^*$ results in considerable performance deterioration. 

\paragraph{Impact on Retrained Constructive NCO}
Given that inconsistencies between training and inference strategies may introduce a deviation between the model's learned perception and its actual decision-making process, we further retrain three reduced variants of LEHD with different SSR configurations. As illustrated in \cref{table:ablation_effects_lehd}, while reducing the search space significantly reduces inference time, employing an excessively small $k$ based solely on pairwise distances results in substantial performance deterioration.

\begin{table}[t]
\centering
\caption{Effects of distance-based SSR in retrained models. All models use identical training settings as the original LEHD.}
%\scalebox{0.9}
\resizebox{\columnwidth}{!}{
\begin{tabular}{l |  cc | cc | cc |  cc }
\toprule[0.5mm]
\multirow{2}{*}{Method }& \multicolumn{2}{c|}{TSP100}& \multicolumn{2}{c|}{TSP200} & \multicolumn{2}{c|}{TSP500}&\multicolumn{2}{c}{TSP1000} \\ 
 & Gap & Time& Gap & Time& Gap & Time& Gap & Time\\
\midrule
LKH3&0.00\% & 56m&  0.00\%  &4m & 0.00\%   & 32m &0.00\% &8.2h \\
\midrule
LEHD w/ $k=20$ &3.10\% &\cellcolor[HTML]{D0CECE}\textbf{10.7s} & 5.21\% & \cellcolor[HTML]{D0CECE}\textbf{1.0s}& 7.95\% &  \cellcolor[HTML]{D0CECE}\textbf{2.1s} &11.09\%  &\cellcolor[HTML]{D0CECE}\textbf{3.7s}\\
LEHD w/ $k=30$ &1.71\% &15s & 3.03\% & 1.1s& 6.01\% & 2.2s  &9.44\% &  4s\\ 
LEHD w/ $k=50$ &0.87\% & 18.7s& 1.74\% & 1.28s& 4.03\%  & 2.6s &7.41\%  & 4.9s\\ 
LEHD w/o SSR &\cellcolor[HTML]{D0CECE}\textbf{0.58\%} & 24s& \cellcolor[HTML]{D0CECE}\textbf{0.86\%} & 3s& \cellcolor[HTML]{D0CECE}\textbf{1.56\%}& 18s   &\cellcolor[HTML]{D0CECE}\textbf{3.17\% }& 1.6m\\ 
\bottomrule[0.5mm]
\end{tabular}
}
\vspace{-10pt}
\label{table:ablation_effects_lehd}
\end{table}
 
\begin{figure*}[t]
\begin{center}
\centerline{\includegraphics[width=0.98\linewidth]{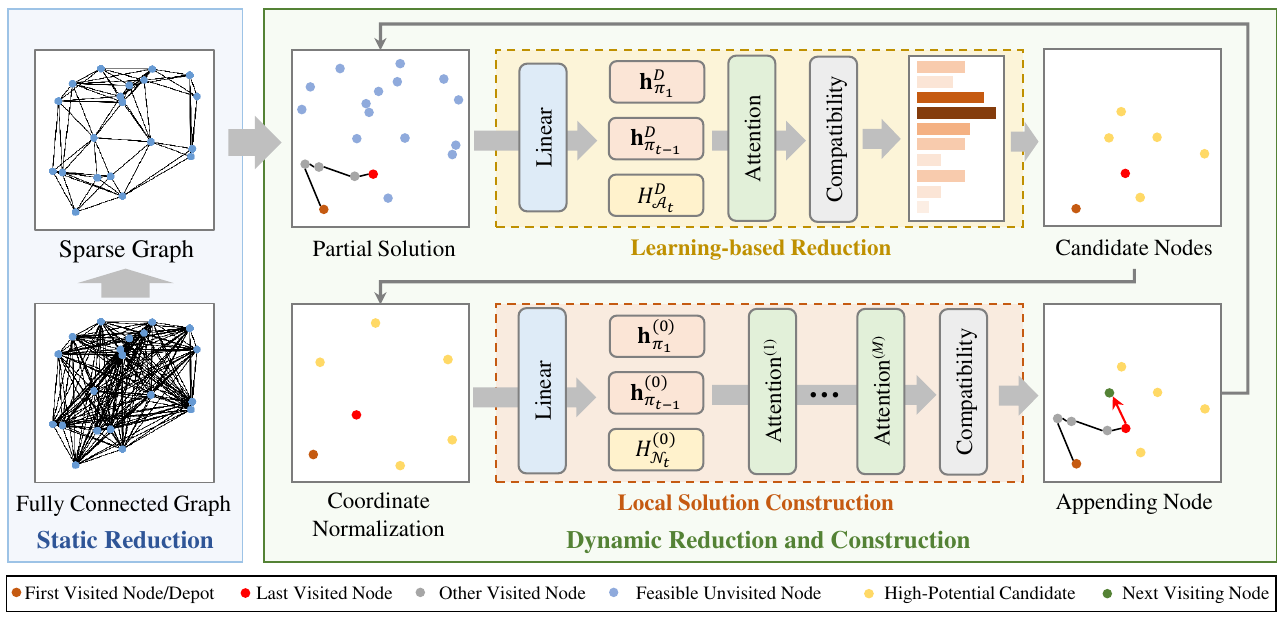}}
\caption{The pipeline of our proposed L2R framework for solving large-scale VRP instances. }
\label{fig:L2R_framework}
\end{center}
\vspace{-18pt}
\end{figure*}
\subsection{Incompatibility with Complex Constraints}
Notably, the limitations of purely distance-based reduction become particularly pronounced when addressing VRP variants with complex constraints, such as the Capacitated VRP (CVRP) and CVRP with Time Windows (CVRPTW). In these problems, the optimal route depends not only on geometric proximity but also on non-spatial constraints such as capacity and time windows. Because distance-based pruning ignores these non‑spatial requirements, it may discard nodes that are geometrically distant yet essential for constraint satisfaction, resulting in significant performance loss.

\subsection{Necessity of Learning-based Reduction}
These findings underscore the potential of SSR while highlighting the critical need for advanced reduction strategies to efficiently handle large-scale problems with small $k$. This is particularly important for addressing large-scale instances with non-uniform distributions or where optimal solutions rely heavily on non-spatial constraints rather than simple geometric proximity.

The powerful learning and knowledge extraction capabilities of deep neural networks offer a viable path to address this need, which enables effective data-driven ranking and selection of high-potential nodes. To balance solution quality and efficiency on large-scale instances, several critical improvements are required: 
\begin{itemize}
    \item \textbf{Lightweight Reduction Model:} To select the best candidates, we need to assess the potential of each feasible node with the current state at each step, which is prohibitively expensive for large-scale problems (i.e., $N \ge 10K$). An efficient, lightweight architecture is therefore essential for dynamic candidate prioritization.
    \item \textbf{Integration of Distance Priors:} Relying solely on a lightweight network to identify promising candidates within a vast search space is inherently difficult. Since distance information is a strong prior, highly correlated with most VRP objectives, integrating learned features with distance priors can reduce the model's learning burden and effectively harness the problem's inherent geometric information.
    \item \textbf{Static Reduction Preprocessing:} Although purely distance-based SSR has limitations, an initial static pruning step that removes clearly unpromising long-distance edges remains a valid strategy for enhancing the computational efficiency of the reduction model.
    \item \textbf{Powerful Local Construction Model:} Given the irreversible nature of sequential node selection, even with a reduced search space, a powerful local solution construction model remains crucial to select the next node to visit at each construction step.
\end{itemize}

\section{Learning to Reduce (L2R)} 
\label{sec:L2R}
In this section, we propose \textit{Learning to Reduce (L2R)}, a novel learning-based dynamic SSR framework to address the limitations of existing SSR methods. As illustrated in \cref{fig:L2R_framework}, our framework introduces three complementary stages: 1) static reduction, 2) learning-based reduction, and 3) local solution construction, each detailed in the following subsections.

\subsection{Static Reduction} 
The static reduction stage initiates our framework by pruning the original fully connected graph $G = (\Omega, E)$ into a sparse topology $G' = (\Omega, E')$. For each node $i \in \Omega$, pairwise Euclidean distances $\{d_{ij}\}_{j=1}^{N}$ are computed, and edges connecting to nodes within the farthest $\gamma$-percentile are removed. Formally, it can be defined as
\begin{equation}
E' = \bigcup_{i \in \Omega} \left\{ e_{ij} \mid \mathrm{rank}(d_{ij}) \leq (1 - \gamma)|\Omega| \right\}, \quad \gamma \in [0,1]
\end{equation}
where $\mathrm{rank}(d_{ij})$ denotes the ascending order of distances from node $v_i$ (i.e., $\mathrm{rank}(d_{ij})=1$ for the closest neighbor). Based on empirical analysis across diverse scales and node distributions, we set $\gamma = 10\%$ as the threshold in this work, which efficiently reduces computational overhead in subsequent stages without compromising solution optimality. 

\subsection{Learning-based Reduction} 
While the static reduction stage removes the most distant edges, the pruned graph $G'$ still contains many non-optimal edge connections, leaving computational costs prohibitive on large-scale instances. At the same time, as discussed in the previous section, further reducing the search space solely based on geometric distance would severely degrade solution quality. To address this challenge, we develop a learning-based model to dynamically evaluate the potential of feasible nodes and adaptively prune the search space at each construction step. For efficiency, the model employs a lightweight architecture with only an embedding layer followed by an attention layer, as detailed below.

\paragraph{Embedding Layer}
Given a VRP instance with an optional depot indexed by $0$ and $n$ customers indexed by $\{1, 2, \ldots, n\}$, it can be represented as $S = \{\mathbf{s}_i\}_{i=0}^n$. In VRP (e.g., CVRP), each node $\mathbf{s}_i$ contains spatial node coordinates $\{x_{i},y_{i}\}$ and problem-specific attributes $\bm{\omega}_{i}$. Specifically, $\bm{\omega}_{i}$ consists of node demand (common to CVRP variants), which is further augmented with time window information (i.e., earliest and latest arrival time) for CVRPTW. These features are first projected into a $d$-dimensional embedding through a shared linear transformation for each node:
\begin{equation}
\mathbf{h}_i^{D} = W^{(e)}[x_{i},\ y_{i},\ \bm{\omega}_{i}] + \mathbf{b}^{(e)}, \  i=0,1,2,\dots,n,
\end{equation}
where $W^{(e)} \in \mathbb{R}^{(2+|\bm{\omega}_{i}|) \times d}$ and $\mathbf{b}^{(e)} \in \mathbb{R}^d$ are learnable parameters. In other words, we obtain a set of embeddings $H^D = \{\mathbf{h}_i^{D}\}_{i=0}^{n} \in \mathbb{R}^{(n+1) \times d}$ for all nodes in the instance $S$. Let $\mathcal{A}_{t}$ denote the set of feasible nodes at step $t$, comprising unvisited nodes with a valid edge in $E^{\prime}$ connected to the current node that satisfy all validity constraints (e.g., capacity and time windows). The embeddings of $\mathcal{A}_{t}$ are denoted by $H_{\mathcal{A}_{t}}^{D} =\{ \mathbf{h}_i^{D}|i\in \mathcal{A}_{t}\}\in \mathbb{R}^{ |\mathcal{A}_{t}| \times d}$. 

\paragraph{Attention Layer}
At the $t$-th step with partial solution $(\pi_1,\dots,\pi_{{t-1}})$, we adopt the setting from \citet{kool2019attention} to represent the current partial solution using the initial node embedding $\mathbf{h}_{\pi_1}^D$ and latest node embedding $\mathbf{h}_{\pi_{t-1}}^D$. Following \citet{kwon2020pomo}, the context embedding of the current partial solution is defined as
\begin{equation}
% \mathbf{h}_{C_{D}}^{t} = W_{\text{first}}\mathbf{h}_{\pi_{1}}^{D} + W_{\text{last}}\mathbf{h}_{\pi_{t-1}}^{D},
\mathbf{h}_{C_{D}}^{t} = \begin{cases}
[\mathbf{h}_{\pi_{t-1}}^{D},Q_{\text{remain}}]W_{C_{D}}  & \text{if} \ Q_{\text{remain}} \neq \emptyset \\
   [\mathbf{h}_{\pi_{1}}^{D},\mathbf{h}_{\pi_{t-1}}^{D}]W_{C_{D}}   &\text{otherwise}, 
    \end{cases}
    \label{eq:dynamic_c}
\end{equation}
where $[\cdot,\cdot]$ is the horizontal concatenation operator, $W_{C_{D}}$ is a linear projection matrix, and $Q_{\text{remain}}$ represents the current remaining load for CVRP variants. 

To calculate potential scores for all feasible nodes in $\mathcal{A}_{t}$, we process the context embedding $\mathbf{h}_{C_D}^t$ and the node embeddings $H_{\mathcal{A}_t}^D$ using an attention mechanism. First, the node embeddings are projected into key-value pairs:
\begin{equation}
K_{\mathcal{A}_t}^D = W^K H_{\mathcal{A}_t}^D, \quad V_{\mathcal{A}_t}^D = W^V H_{\mathcal{A}_t}^D
\label{eq:qkv_dynamicstage}
\end{equation}
where $W^K, W^V \in \mathbb{R}^{d \times d}$ are learnable projection matrices. The context embedding $\mathbf{h}_{C_D}^t$ then interacts with these projections through the attention operator:
\begin{equation}
\hat{\mathbf{h}}_{C_D}^t = \mathrm{Attention}\left(\mathbf{h}_{C_D}^t, K_{\mathcal{A}_t}^D, V_{\mathcal{A}_t}^D\right)
\label{eq:dynamic_aafm}
\end{equation}

\paragraph{Compatibility Calculation}
Finally, similar to previous work~\citep{zhou2024icam}, we compute the compatibilities $\mathbf{u}^{R}=\{u^{R}_{t,i}| i \in \mathcal{A}_{t}\}$:
\begin{equation}
    u^{R}_{t,i} = 
    \begin{cases}
    \xi  \cdot \text{tanh}\left(\frac{\hat{\mathbf{h}}_{C_D}^{t}(\mathbf{h}_{i}^{D})^\mathrm{T}}{\sqrt{d_k}} +a_{t-1,i}^{R}\right)   & \text{if} \ i \in \mathcal{A}_{t} \\
    -\infty  &\text{otherwise}
\end{cases},
    \label{eq:u_calculated_reduction}
\end{equation}
where $\xi$ is the clipping parameter, $d_k$ is the dimension for matrix $K_{\mathcal{A}_{t}}^{D}$. Following \citet{zhou2024icam}, the adaptation bias between each node $i \in \mathcal{A}_{t}$ and the current node $\pi_{t-1}$ is defined as $a_{t-1,i}^{R} = -\alpha\cdot\log_2{N}\cdot d_{t-1,i}$, where $d_{t-1,i}$ is their distance, $N$ is the problem size, and $\alpha > 0$ is a learnable parameter (defaulting to 1). This formulation can better capture diverse geometric patterns. Finally, we compute the potential scores $\mathbf{o} = \mathrm{softmax}(\mathbf{u}^{R})$ and retain the top-$k$ nodes with the highest scores to construct the candidate set $\mathcal{N}_{t}$, thereby performing a learning-based dynamic SSR.

By adaptively integrating static distance priors with learned context-aware features, our reduction model generates potential scores $\mathbf{o}$ for the set of feasible nodes $\mathcal{A}_{t}$. This enables it to dynamically balance physical proximity against real-time problem states (e.g., remaining load), effectively solving complex VRPs where optimal solutions depend not only on geometric distances but also on non-spatial constraints. The empirical validity of this design is demonstrated by our results on CVRP and CVRPTW instances, as detailed in Table \ref{table:uniform_tsp_cvrp} and Table \ref{table:cvrptw}.

\subsection{Local Solution Construction} 
We develop a learning-based local solution construction model to select a final node from the candidate set $\mathcal{N}_{t}$ to extend the partial solution at construction step $t$. First, we introduce a coordinate normalization operation to ensure that each extracted sub-graph $G_{\text{sub}}^{'}$ adheres to a similar distribution, following \citep{fu2021attgcn_mcts,fang2024invit}. Taking TSP as an example, as in previous work~\citep{kwon2020pomo,luo2023lehd}, we use the initial node $\pi_{1}$ and the last-visited node $\pi_{t-1}$ to represent the current partial solution. First of all, the embedding of each node can be obtained by a shared linear transformation $\mathbf{h}_i^{(0)} = W^{(0)}\mathbf{s}_i + \mathbf{b}^{(0)}, \quad \forall i \in \{\pi_{1}, \pi_{t-1}\} \cup \mathcal{N}_{t}$, where $W^{(0)}$ and $b^{(0)}$ are learnable parameters. Then we obtain the embedding of the partial graph: $\widetilde{H}^{(0)}=[W_{1}\mathbf{h}_{\pi_1}^{(0)}, \ W_{2}\mathbf{h}_{\pi_{t-1}}^{(0)},\  H_{\mathcal{N}_{t}}^{(0)}]\in \mathbb{R}^{(2+|\mathcal{N}_{t}|) \times d}$, where $[\cdot,\cdot]$ denotes the vertical concatenation operator, $H_{\mathcal{N}_{t}}^{(0)} =\{ \mathbf{h}_i^{(0)}|i\in \mathcal{N}_{t}\}\in \mathbb{R}^{|\mathcal{N}_{t}| \times d}$ are node embeddings for $\mathcal{N}_{t}$, $W_{1}\in \mathbb{R}^{d \times d}$ and $W_{2}\in \mathbb{R}^{d \times d}$ are two learnable matrices. This embedding is the initial input to a sequence of attention layers. 

In the proposed model, each attention layer includes an attention sub-layer and a Feed-Forward (FF) sub-layer, both equipped with Layer Normalization~\citep{ba2016layernorm} and skip-connection~\citep{he2016skip} (see Appendix \ref{append:local_attention_layer} for details). After $M$ attention layers, $\widetilde{H}^{(M)}=[\mathbf{h}_{\pi_1}^{(M)}, \ \mathbf{h}_{\pi_{t-1}}^{(M)},\  H_{\mathcal{N}_{t}}^{(M)}]\in \mathbb{R}^{(2+|\mathcal{N}_{t}|) \times d}$ contains advanced embeddings of the initial, last, and $k$ candidate nodes.

Finally, a separate compatibility module computes the logits $\mathbf{u}^{L}=\{u^{L}_{t,i}| i \in \mathcal{N}_{t}\}$ for all nodes in $\mathcal{N}_{t}$. The probabilities can then be calculated as $\mathbf{p} = \mathrm{softmax}(\mathbf{u}^{L})$. See Appendix~\ref{append:local_construction_model} for more details.

\subsection{Training} 
Let $\bm{\theta} =\{\bm{\theta}_{R}, \ \bm{\theta}_{L}\}$ denote learnable parameters, where $\bm{\theta}_{R}$ and $\ \bm{\theta}_{L}$ are the reduction and construction models, respectively. At each step $t$, $\bm{\theta}_{L}$ samples a node $\pi_{t}$ from the distribution $\mathbf{p}=\{p_{i}|i \in \mathcal{N}_{t}\}$ and appends it to the partial solution. After $m$ steps, a complete solution $\pi = (\pi_{1},\dots,\pi_{m})$ for an instance $S$ is constructed. Since $\bm{\theta}_{R}$ is responsible for prioritizing feasible nodes to filter the candidate set $\mathcal{N}_t$, rather than directly selecting the next-visit node, we train both $\bm{\theta}_R$ and $\bm{\theta}_L$ using the same reward $\mathcal{R}(\pi \mid S, \bm{\theta})$, defined as the negative tour length of $\pi$ for $S$. We define $\tau=(\tau_{1},\dots,\tau_{m})$ as the best candidate node sampled by $\bm{\theta}_{R}$. Following~\citep{kool2019attention,kwon2020pomo}, $\bm{\theta}_{R}$ and $\bm{\theta}_{L}$ are jointly trained using the REINFORCE algorithm~\citep{williams1992reinforce}: 
\begin{equation}
\begin{aligned}
    \mathcal{L}_{\mathrm{Joint}} (\bm{\theta}) &= \mathcal{L}_{R} (\bm{\theta}_{R}) + \mathcal{L}_{L} (\bm{\theta}_{L}), \\
    \nabla_{\bm{\theta}_{R}} \mathcal{L}_{R} (\bm{\theta}_{R}) &= \mathbb{E}_{o(\tau| S,{\bm{\theta}_{R}})} \left[\left(\mathcal{R}(\pi| S,{\bm{\theta}})-b(S)\right) \nabla_{\bm{\theta}_{R}} \log o(\tau| S)\right], \\
    \nabla_{\bm{\theta}_{L}} \mathcal{L}_{L} (\bm{\theta}_{L}) &= \mathbb{E}_{p(\pi| S,{\bm{\theta}_{L}})} \left[\left(\mathcal{R}(\pi| S,{\bm{\theta}})-b(S)\right) \nabla_{\bm{\theta}_{L}} \log p(\pi| S)\right],
\end{aligned}
\label{eq:rl_loss_joint}
\end{equation}
where $o(\mathbf{\tau}\mid S)=\prod_{t=2}^{n}o(\tau_t\mid S,\pi_{1:t-1})$, $p(\mathbf{\pi}\mid S)=\prod_{t=2}^{n}p(\pi_t\mid S,\pi_{1:t-1})$, and $b(S)$ is the greedy rollout baseline~\citep{kool2019attention}.

\section{Experiments}
\label{sec:experiments}
In this section, we comprehensively evaluate L2R against both classical and learning-based solvers on large-scale TSP, CVRP, and CVRPTW instances. Notably, the model for each problem is trained exclusively on uniformly sampled 100-node instances. Aligned with our core contribution on advancing generalization and scalability in NCO, we focus our assessment on two key aspects: (1) scalability to problem sizes up to 10 million nodes and (2) robustness to diverse node distributions and benchmark datasets.  All experiments are conducted on a single NVIDIA GeForce RTX 3090 GPU. 

\begin{table*}[t]
\caption{Comparison of TSP and CVRP instances with uniform distribution (1K $\leq N\leq$ 100K).}
\centering
\resizebox{0.96\textwidth}{!}{
\begin{tabular}{l | cc | cc | cc | cc | cc}
\toprule[0.5mm]
\multirow{2}{*}{Method}& \multicolumn{2}{c|}{TSP1K}   & \multicolumn{2}{c|}{TSP5K} & \multicolumn{2}{c|}{TSP10K}& \multicolumn{2}{c|}{TSP50K} & \multicolumn{2}{c}{TSP100K} \\ 
& Obj. (Gap) & Time  & Obj. (Gap) & Time & Obj. (Gap) & Time & Obj. (Gap) & Time & Obj. (Gap) & Time     \\ 
\midrule
 LKH3   & 23.12 (0.00\%) & 1.7m &  50.97 (0.00\%) & 12m & 71.78 (0.00\%) & 33m &  159.93 (0.00\%)  & 10h & 225.99 (0.00\%)  & 25h\\
 Concorde & 23.12 (0.00\%) & 1m &  50.95 (-0.05\%) & 31m & 72.00 (0.15\%) & 1.4h   & N/A  & N/A & N/A  & N/A \\
\midrule
H-TSP & 24.66 (6.66\%) & 48s   & 55.16 (8.21\%) & 1.2m  & 77.75 (8.38\%) & 2.2m  & \multicolumn{2}{c|}{OOM} & \multicolumn{2}{c}{OOM} \\
GLOP   & 23.78 (2.85\%) & 10.2s  & 53.15 (4.26\%) & 1.0m & 75.04 (4.39\%) & 1.9m   &  168.09 (5.10\%) & 1.5m & 237.61 (5.14\%)  &  3.9m\\
\midrule
 LEHD greedy & 23.84 (3.11\%) & 0.8s &  58.85 (15.46\%)  & 1.5m &  91.33 (27.24\%)  & 11.7m & \multicolumn{2}{c|}{OOM} & \multicolumn{2}{c}{OOM}    \\
 LEHD RRC1,000    &  \cellcolor[HTML]{D0CECE}\textbf{23.29 (0.72\%)} & 3.3m & 54.43 (6.79\%) & 8.6m & 80.90 (12.5\%) & 18.6m &  \multicolumn{2}{c|}{OOM}  &   \multicolumn{2}{c}{OOM} \\
 BQ greedy & 23.65 (2.30\%) & 0.9s &  58.27 (14.31\%)  & 22.5s &  89.73 (25.02\%)  & 1.0m & \multicolumn{2}{c|}{OOM} & \multicolumn{2}{c}{OOM}   \\
 BQ bs16   & 23.43 (1.37\%) & 13s  & 58.27 (10.7\%) & 24s & \multicolumn{2}{c|}{OOM}  & \multicolumn{2}{c|}{OOM}   &   \multicolumn{2}{c}{OOM} \\
\midrule
POMO aug$\times$8    & 32.51 (40.6\%) & 4.1s & 87.72 (72.1\%)  & 8.6m  &  \multicolumn{2}{c|}{OOM} &  \multicolumn{2}{c|}{OOM}    &  \multicolumn{2}{c}{OOM} \\
ELG aug$\times$8   &  25.74 (11.33\%) & 0.8s  & 60.19 (18.08\%) & 21s &    \multicolumn{2}{c|}{OOM}  & \multicolumn{2}{c|}{OOM}    &  \multicolumn{2}{c}{OOM}\\
DGL greedy & 24.41 (5.58\%) & 0.1s & 54.63 (7.18\%) & 2.3s &77.24 (7.61\%) & 4.3s & \multicolumn{2}{c|}{OOM} & \multicolumn{2}{c}{OOM} \\
INViT-3V greedy     & 24.67 (6.71\%) & 0.4s & 54.46 (6.84\%) & 12.7s & 76.87 (7.09\%) & 34.9s & 171.42 (7.18\%) & 4.9m & 242.26 (7.20\%) & 18.8m \\
\midrule
L2R greedy    &  24.16 (4.49\%) & 0.05s &   53.36 (4.69\%) & 1.8s &   75.24 (4.82\%) & 4.1s  &   167.70 (4.86\% ) & 35.5s &  236.81 (4.79\%) &  1.8m \\
L2R PRC100    & 23.62 (2.18\%) & 2.5s &   52.41 (2.82\%) & 20.1s &  73.95 (3.03\%) & 26.3s  &  165.16 (3.27\% ) & 1.8m & 234.36 (3.70\%) &  3.2m \\
L2R PRC1,000   & 23.52 (1.72\%)  & 24.9s &  \cellcolor[HTML]{D0CECE}\textbf{52.20 (2.40\%)} & 3.1m &  \cellcolor[HTML]{D0CECE}\textbf{73.66 (2.62\%)} & 3.8m  &\cellcolor[HTML]{D0CECE}\textbf{ 164.41 (2.80\% )} &  12.4m  & \cellcolor[HTML]{D0CECE}\textbf{232.77 (3.00\%)} &  15.5m\\
\midrule
\midrule
\multirow{2}[2]{*}{Method} & \multicolumn{2}{c|}{CVRP1K} & \multicolumn{2}{c|}{CVRP5K} & \multicolumn{2}{c|}{CVRP10K} & \multicolumn{2}{c|}{CVRP50K} & \multicolumn{2}{c}{CVRP100K} \\
          & Obj.(Gap) & Time  & Obj.(Gap) & Time  & Obj.(Gap) & Time  & Obj.(Gap) & Time  & Obj.(Gap) & Time \\
\midrule
HGS   & 41.2 (0.00\%) & 5m    & 126.2 (0.00\%) & 5m    & 227.2 (0.00\%) & 5m    & 1081.0 (0.00\%) & 4h    & 2087.5 (0.00\%) & 6h \\
\midrule
TAM-LKH3* & 46.3 (12.38\%) & 1.8s  & 144.6 (14.58\%) &17s &   \multicolumn{2}{c|}{$-$}   &   \multicolumn{2}{c|}{$-$} &   \multicolumn{2}{c}{$-$}\\
GLOP-G (LKH-3) & 45.9 (11.4\%) & 1.1s  & 140.6(11.4\%) & 4.0s  & 256.4 (12.85\%) & 6.2s  & \multicolumn{2}{c|}{OOM} & \multicolumn{2}{c}{OOM} \\
\midrule
LEHD greedy & 44 (6.80\%) & 0.8s  & 138.2 (9.51\%) & 1.4m  & 256.3 (12.81\%) & 12m   & \multicolumn{2}{c|}{OOM} & \multicolumn{2}{c}{OOM} \\
LEHD RRC1,000 & \cellcolor[HTML]{D0CECE}\textbf{42.4 (2.91\%)} & 3.4m  & 132.7 (5.15\%) & 10m   & 243.8 (7.31\%) & 51.6m & \multicolumn{2}{c|}{OOM} & \multicolumn{2}{c}{OOM} \\
BQ greedy & 44.2 (7.28\%) & 1s    & 139.9 (10.86\%) & 18.5s & 262.2 (15.40\%) & 2m    & \multicolumn{2}{c|}{OOM} & \multicolumn{2}{c}{OOM} \\
BQ bs16 & 43.1 (4.61\%) & 14s   & 136.4 (8.08\%) & 2.4m  & \multicolumn{2}{c|}{OOM} & \multicolumn{2}{c|}{OOM} & \multicolumn{2}{c}{OOM} \\
\midrule
POMO aug $\times$ 8 & 101 (145.15\%) & 4.6s  & 632.9 (401.51\%) & 11m   & \multicolumn{2}{c|}{OOM} & \multicolumn{2}{c|}{OOM} & \multicolumn{2}{c}{OOM} \\
ELG aug $\times$ 8 & 46.4 (12.62\%) & 10.3s & \multicolumn{2}{c|}{OOM} & \multicolumn{2}{c|}{OOM} & \multicolumn{2}{c|}{OOM} & \multicolumn{2}{c}{OOM} \\
DGL greedy &51.0 (23.82\%) & 0.1s &145.1 (14.99\%) & 0.5s &247.4 (8.89\%) & 4.5s & \multicolumn{2}{c|}{OOM} & \multicolumn{2}{c}{OOM} \\
INViT-3V greedy & 48.3 (17.23\%) & 1s   & 146.2 (15.85\%) & 7s  & 262.5 (15.54\%) & 1.2m  & 1331.1 (23.1\%) & 5.6m  & 2683.4 (28.55\%) & 22m \\
\midrule
L2R greedy& 45.8 (11.37\%) &0.1s  & 136.0 (7.72\%) & 0.5s  & 236.7 (4.17\%) & 4.4s  &  1075.0 (-0.56\%) & 37s & 2055.1 (-1.55\%) & 2m \\
L2R PRC100  &44.7 (8.53\%)& 4.0s & 132.7 (5.14\%)& 14.4s & 233.2 (2.62\%) &39.9s & 1072.6 (-0.78\%) & 3.5m&2051.5 (-1.73\%) &7m\\
L2R PRC1,000  &44.3 (7.42\%) &40.3s & \cellcolor[HTML]{D0CECE}\textbf{130.6 (3.52\%)} &2.3m  & \cellcolor[HTML]{D0CECE}\textbf{230.1 (1.26\%)}& 6.0m & \cellcolor[HTML]{D0CECE}\textbf{1070.5 (-0.98\%)} & 33m   & \cellcolor[HTML]{D0CECE}\textbf{2050.9 (-1.75\%)} & 1.1h \\
\bottomrule[0.5mm]
\end{tabular}
}
\label{table:uniform_tsp_cvrp}
\end{table*}

\begin{table}[htbp]
  \centering
  \caption{Comparison on CVRPTW with 16 instances per scale. }
  \resizebox{0.46\textwidth}{!}{
  \begin{threeparttable}
    \begin{tabular}{c|cc|cc|cc}
    \toprule[0.5mm]
    \multirow{2}{*}{Method} & \multicolumn{2}{c|}{CVRPTW1K} & \multicolumn{2}{c|}{CVRPTW5K} & \multicolumn{2}{c}{CVRPTW10K} \\
          & Gap   & Time  & Gap   & Time  & Gap   & Time \\
    \midrule
    HGS-PyVRP & 0.00\% & 1.3m   & 0.00\% & 12.4m  & 0.00\% & 18.8m \\
    \midrule
    MTPOMO$\dagger$ & 43.94\% & 1.2s  & 69.11\% & 32.7s & 82.58\% & 121.4s \\
    MVMoE$\dagger$ & 49.16\% & 1.3s  & 151.53\% & 36.3s & 188.38\% & 130.4s \\
    ReLD-MTL$\dagger$ & 15.93\% & 1.1s  & 22.40\% & 30.8s & 25.99\% & 113.8s \\
    \midrule
    L2R greedy & \cellcolor[HTML]{D0CECE}\textbf{9.39\%} & \cellcolor[HTML]{D0CECE}\textbf{0.5s} & \cellcolor[HTML]{D0CECE}\textbf{7.95\%} & \cellcolor[HTML]{D0CECE}\textbf{2.5s} & \cellcolor[HTML]{D0CECE}\textbf{8.83\%}      & \cellcolor[HTML]{D0CECE}\textbf{5.2s} \\
    \bottomrule[0.5mm]
    \end{tabular}%
    \begin{tablenotes}
\item[$\dagger$] The instance augmentation is removed to prevent OOM.
\end{tablenotes}
\end{threeparttable}
}
  \label{table:cvrptw}%
  \vspace{-18pt}
\end{table}%

\begin{table}[t]
  \centering
  \caption{Comparison with SIL~\citep{luo2024SIL} using large-scale training.}
  \resizebox{0.48\textwidth}{!}{
    \begin{tabular}{c|cc|cc}
    \toprule[0.5mm]
    \multirow{2}[1]{*}{Method}  & \multicolumn{2}{c|}{TSP50K} & \multicolumn{2}{c}{TSP100K} \\
           & Obj.(Gap) & Time  & Obj.(Gap) & Time \\
    \midrule
    LKH3     & 159.93 (0.00\%) & 10h   & 225.99 (0.00\%) & 25h \\
    \midrule
    SIL-1K greedy  & 178.99 (11.92\%) & 8.6m  & 266.63 (17.98\%) & 36.2m \\
    L2R-100 greedy    & \cellcolor[HTML]{D0CECE}\textbf{167.70 (4.86\%)} & \cellcolor[HTML]{D0CECE}\textbf{35.5s} & \cellcolor[HTML]{D0CECE}\textbf{236.81 (4.79\%)} & \cellcolor[HTML]{D0CECE}\textbf{1.8m} \\
    \midrule
    SIL-1K PRC1000   & 167.01 (4.43\%) & 1.6h  & 241.68 (6.94\%) & 3.0h \\
    L2R-100 PRC1000    & \cellcolor[HTML]{D0CECE}\textbf{ 164.41 (2.80\%)} & \cellcolor[HTML]{D0CECE}\textbf{ 12.4m} & \cellcolor[HTML]{D0CECE}\textbf{232.77 (3.00\%)} & \cellcolor[HTML]{D0CECE}\textbf{15.5m} \\
    \midrule
    \midrule
     \multirow{2}[1]{*}{Method} & \multicolumn{2}{c|}{CVRP50K} & \multicolumn{2}{c}{CVRP100K} \\
           & Obj.(Gap) & Time  & Obj.(Gap) & Time \\
    \midrule
    HGS      & 1081.00 (0.00\%) & 4h    & 2087.51 (0.00\%) & 6h \\
    \midrule
    SIL-1K greedy   & 1108.28 (2.52\%) & 9.2m  & 2153.78 (3.17\%) & 45.8m \\
    L2R-100 greedy   & \cellcolor[HTML]{D0CECE}\textbf{1074.98 (-0.56\%)} & \cellcolor[HTML]{D0CECE}\textbf{37s} & \cellcolor[HTML]{D0CECE}\textbf{2055.14 (-1.55\%)} & \cellcolor[HTML]{D0CECE}\textbf{2m} \\
    \midrule
    SIL-1K PRC1000   & 1094.63 (1.26\%) & 1.6h  & 2140.02 (2.52\%) & 3.5h \\
    L2R-100 PRC1000   & \cellcolor[HTML]{D0CECE}\textbf{1070.49 (-0.98\%)} & \cellcolor[HTML]{D0CECE}\textbf{33m} & \cellcolor[HTML]{D0CECE}\textbf{2050.91 (-1.75\%)} & \cellcolor[HTML]{D0CECE}\textbf{1.1h} \\
    \bottomrule[0.5mm]
    \end{tabular}%
    }
    \vspace{-5pt}
  \label{table:compare_SIL}%
\end{table}%

\begin{table*}[t]
\centering
\caption{Comparison on cross-distribution generalization with $20$ instances per dataset.}
\resizebox{0.96\textwidth}{!}{
\begin{threeparttable}
\begin{tabular}{l |  cc | cc |  cc |  cc | cc |  cc}
\toprule[0.5mm]
\multirow{2}{*}{Method} & \multicolumn{2}{c|}{TSP5K, Cluster} & \multicolumn{2}{c|}{TSP5K, Explosion}& \multicolumn{2}{c|}{TSP5K, Implosion} & \multicolumn{2}{c|}{CVRP5K, Cluster} & \multicolumn{2}{c|}{CVRP5K, Explosion}& \multicolumn{2}{c}{CVRP5K, Implosion} \\ 
 &Gap & Time  &Gap & Time & Gap & Time  &Gap & Time  &Gap & Time & Gap & Time \\
\midrule
Near-optimal &  0.00\% & $-$ & 0.00\% &   $-$   &0.00\% & $-$ &  0.00\% & $-$ & 0.00\% &   $-$   &0.00\% & $-$\\
\midrule
ELG$\ddagger$ & 22.83\% & 1.6m & 20.71\% & 1.6m   &17.55\% &  1.6m &18.14\% & 2.4m &13.21\% &  2.2m  &7.50\% & 2.2m  \\
Omni\_VRP$\ddagger$ & 54.53\% & 1.1m & 51.09\% &  1.1m  &50.20\% & 1.1m  & 22.05\% & 1.3m & 33.09\% &  1.4m  &40.20\% & 1.3m  \\
INViT-3V greedy  & 8.20\% & 11.3s &11.48\% &  11.3s  &8.52\% & 11.3s   &9.05\% & 29.4s & 8.44\% & 29.3s   & 8.77\% & 23.2s  \\
\midrule
L2R greedy  & 6.14\% & 1.5s & 9.54\% & 1.5s   &6.17\% & 1.5s  & 2.65\% & 1.8s & 4.38\% &  1.8s  &3.15\% & 1.8s  \\
L2R PRC1,000  & \cellcolor[HTML]{D0CECE}\textbf{3.16\%} & 2.6m & \cellcolor[HTML]{D0CECE}\textbf{3.52\%} & 2.6m   &\cellcolor[HTML]{D0CECE}\textbf{2.87\%} & 2.6m  & \cellcolor[HTML]{D0CECE}\textbf{0.99\%} & 4.2m & \cellcolor[HTML]{D0CECE}\textbf{1.14\%} &  4.2m  &\cellcolor[HTML]{D0CECE}\textbf{0.73\%} & 4.2m  \\

\bottomrule[0.5mm]
\end{tabular}
\begin{tablenotes}
\item[$\ddagger$] The instance augmentation technique is not employed for comparable methods to prevent methods from exceeding memory limits. 
\end{tablenotes}
\end{threeparttable}
}
\label{table:cross_distribution}
\vspace{-5pt}
\end{table*}

\begin{table*}[t]
\centering
\caption{Comparison on large-scale TSPLib~\citep{reinelt1991tsplib} ($1,000 \leq N \leq 85,900$) and CVRPLib~\citep{arnold2019cvrplib_xxl} instances ($3,000 \leq N \leq 30,000$).}
\resizebox{0.96\textwidth}{!}{
\begin{threeparttable}
\begin{tabular}{ l | c | c| c | c| c | c| c | c}
\toprule[0.5mm]
 & \multicolumn{4}{c|}{TSPLib} & \multicolumn{4}{c}{CVRPLib-XXL} \\
\midrule
\multirow{2}{*}{Method} & 1K  $\leq N \leq$ 5K & 5K \textless $N \leq$ 100K & All & \multirow{2}{*}{Solved\#} & 3K $\leq N \leq$ 7K & 7K \textless $N \leq$ 30K & All & \multirow{2}{*}{Solved\#}\\
 & (23 instances)&(10 instances) & (33 instances)  &  & (4 instances)&(6 instances) & (10 instances) &\\
\midrule
GLOP & 6.16\%& 7.69\%& 6.62\% & 33/33 &17.07\% & 21.32\%& 19.62\% &10/10\\
TAM-LKH3* & $-$& $-$& $-$& $-$ & 20.44\% & $-$& $-$& $-$\\
BQ bs16   & 10.65\% & 30.58\%\textdagger &$-$ & 26/33 & 20.20\% & OOM &$-$  & 4/10\\
LEHD greedy  & 11.14\% & 39.34\%\textdagger &$-$& 30/33  & 22.22\% & 32.80\%\textdagger &$-$ & 6/10\\
LEHD RRC1000  & 4.00\% & 18.46\%\textdagger &$-$& 30/33& 14.06\% & 21.52\%\textdagger &$-$ & 6/10\\
POMO aug$\times$8 & 62.81\%& OOM& $-$ & 23/33& 331.24\%\textdagger& OOM& $-$ & 2/10\\
ELG aug$\times$8   & 11.34\% & OOM &$-$& 23/33& 16.82\% \textdagger & OOM &  $-$& 2/10\\
INViT-3V greedy   & 11.49\% & 10.03\% &11.05\% &33/33 & 20.74\% & 26.64\% & 24.28\%& 10/10\\
\midrule
L2R greedy   & 9.16\% & 7.36\% &8.61\%&  33/33 & 12.05\% & 11.22\% &11.55\% & 10/10\\
L2R PRC1000  & \cellcolor[HTML]{D0CECE}\textbf{2.80\%} & \cellcolor[HTML]{D0CECE}\textbf{3.96\%} &\cellcolor[HTML]{D0CECE}\textbf{3.15\%}&  33/33 &\cellcolor[HTML]{D0CECE}\textbf{ 7.81\%} & \cellcolor[HTML]{D0CECE}\textbf{ 8.37\%} &\cellcolor[HTML]{D0CECE}\textbf{ 8.14\%} & 10/10\\
\bottomrule[0.5mm]
\end{tabular}
\begin{tablenotes}
\item[\textdagger] Some instances are skipped due to the OOM issue.
\end{tablenotes}
\end{threeparttable}
}
\label{table:TSPLIB_CVRPLIB}
\vspace{-10pt}
\end{table*}
\subsection{Experimental Setup}
\paragraph{Problem Setting} 
For TSP and CVRP, we generate synthetic instances following the methodology outlined in \citep{kool2019attention}. For CVRPTW, we generate instances following MVMoE~\citep{zhou2024mvmoe}. Specifically, for TSP and CVRP, we construct test datasets with uniformly distributed nodes at five scales: 1K, 5K, 10K, 50K, and 100K. For CVRPTW, we limit the test scale to 10K since existing methods consistently run out of memory (OOM) on larger scales, and each has 16 instances. For CVRP(TW), we adhere to the capacity settings in \citet{hou2022tam}. For TSP, following \citep{fu2021attgcn_mcts}, the TSP1K test set consists of 128 instances, while the larger scales each contain 16 instances. For CVRP, each dataset contains 100 instances for scales with $\textless$ 10K nodes~\citep{hou2022tam} and 16 instances for scales with $\geq$ 10K nodes. To evaluate cross-distribution generalization, we test L2R on the TSP/CVRP5K instances obtained from \citep{fang2024invit} and three benchmark datasets: (1) symmetric EUC\_2D instances from TSPLib \citep{reinelt1991tsplib} (1K$\leq N \leq$ 100K); (2) CVRPLib Set-XXL \citep{arnold2019cvrplib_xxl} (3K$\leq N \leq$ 30K), and (3) 8th DIMACS Implementation Challenge~\citep{dimacs} (10K$\leq N \leq$ 10M). 

\paragraph{Model \& Training Setting}
For all experiments, we use an embedding dimension of $128$ and a feed-forward layer dimension of $512$. To enhance geometric pattern recognition in VRPs, we adopt the attention mechanism introduced by \citet{zhou2024icam}. The local construction model employs $6$ attention layers. We set the clipping parameter $\xi = 10$~\citep{kool2019attention}. The hyperparameter $k$ is configured as $20$ for TSP, $50$ for CVRP, and $15$ for CVRPTW.

Our models are exclusively trained on uniformly distributed instances with $100$ nodes to ensure a fair comparison with existing methods. We employ the Adam optimizer~\citep{kingma2014adam} with an initial learning rate of $\eta = 10^{-4}$ and a learning rate decay of $0.98$ per epoch. Training spans $100$ epochs with $2\text{,}500$ batches per epoch. Due to memory constraints, batch sizes differ across problems: $180$ for TSP, $60$ for CVRP, and $128$ for CVRPTW. The same pre-trained model is used in all experimental evaluations for each problem. 

\paragraph{Baseline}
We compare L2R with the following methods: (1)\textbf{Classical Solver}: Concorde~\citep{applegate2006concorde}, LKH3~\citep{LKH3}, HGS~\citep{HGS,wouda2024pyvrp}; (2)\textbf{Constructive NCO}: POMO~\citep{kwon2020pomo}, Omni\_VRP~\citep{zhou2023omni}, ELG~\citep{gao2023elg}, BQ~\citep{drakulic2023bq}, LEHD~\citep{luo2023lehd}, INViT~\citep{fang2024invit}, DGL~\citep{xiao2025dgl}, MTPOMO~\citep{liu2024mtpomo}, MVMoE~\citep{zhou2024mvmoe}, ReLD-MTL~\citep{huang2025reld}, and SIL~\citep{luo2024SIL}; (3)\textbf{Two-Stage NCO}: TAM~\citep{hou2022tam} and GLOP~\citep{ye2023glop}.

\paragraph{Metrics and Inference}
We report the average objective value (Obj.), optimality gap (Gap), and average inference time (Time) for each method. The gap quantifies the discrepancy between the generated solutions and the near-optimal solutions. For most NCO baselines, we execute the official source code using default settings. Results marked with an asterisk (*) are directly obtained from the corresponding papers. Some methods fail to produce feasible solutions within a reasonable time limit (e.g., one week), as indicated by 'N/A'. The notation 'OOM' indicates that the memory consumption exceeds the available memory limits. For L2R, we report two types of results: those obtained by greedy decoding and those derived from Parallel local ReConstruction (PRC) with different numbers of iterations~\citep{luo2024SIL}. The parallel approach shows promising results by effectively trading computing time for improved solution quality. For L2R, initial PRC solutions are generated using greedy decoding. Further details about PRC are available in \citep{luo2024SIL}.
 
\subsection{Performance Evaluation}
\paragraph{Cross-Size Generalization}
We conduct comprehensive experiments on large-scale VRP instances with uniform distribution, and the results are reported in \cref{table:uniform_tsp_cvrp} (TSP/CVRP1K$-$100K) and \cref{table:cvrptw} (CVRPTW1K-10K). Benefiting from our efficient SSR, L2R achieves superior scalability for various VRP variants. While it does not surpass SL-based LEHD on TSP/CVRP1K, the RL-based L2R requires shorter average runtime (e.g., $25\,\text{sec}$ vs. $3.3\,\text{min}$ per TSP1K instance) and outperforms all comparable RL-based methods. For larger-scale TSP and CVRP instances, as well as CVRPTW instances across all scales, L2R consistently achieves the best solution quality among comparable methods while delivering highly efficient inference.

Even compared against classical heuristics that have been refined for decades, L2R demonstrates impressive competitiveness. Most notably, \textbf{even with simple greedy decoding, L2R already surpasses the powerful HGS on CVRP50K and CVRP100K instances}, while achieving a dramatic speedup (e.g., \textbf{2 minutes vs. 6 hours per CVRP100K instance}). To the best of our knowledge, L2R-greedy is the first constructive neural solver to surpass heavily refined HGS on large-scale CVRP instances when \textbf{trained solely on 100-node uniform instances} without any time-consuming post-processing or additional large-scale training. 

These promising results confirm that our proposed L2R can successfully capture complex non-spatial constraints that simple distance-based heuristics cannot, proving the broad applicability beyond simple routing problems.

\paragraph{Comparison against Large-Scale Trained Baseline}
To further highlight the generalization superiority of L2R, we extend our evaluation to include the powerful SIL~\citep{luo2024SIL}, which employs scale-specific training and is fundamentally different from our zero-shot setting. For a challenging comparison, we test L2R-100 (trained on 100 nodes) against SIL-1K (trained on 1,000 nodes) on instances with 50K and 100K nodes. As shown in \cref{table:compare_SIL}, even with a 10$\times$ training scale disadvantage, L2R consistently outperforms SIL in both solution quality and inference efficiency. This outcome highlights the effectiveness of our learning-based SSR. Notably, the greedy version of L2R (L2R-greedy) alone surpasses SIL-PRC1000 on TSP100K and all CVRP instances. These results further confirm that L2R possesses robust generalization capabilities and practical efficiency.

\paragraph{Cross-Distribution Generalization}  
We evaluate the performance of L2R on 5000-node TSP and CVRP instances with three distinct distributions: cluster, explosion, and implosion. All datasets are obtained from INViT~\citep{fang2024invit}, and each contains 20 instances. As shown in \cref{table:cross_distribution}, L2R consistently achieves the best performance among all comparable methods across these distributions, which further highlights the robust generalization capabilities of our L2R. 

\begin{table}[t]
  \centering
  \caption{Experimental results tested on the 8th DIMACS challenge E series TSP instances (10K $\leq N \leq$ 10M).}
  \resizebox{0.98\columnwidth}{!}{
   \begin{tabular}{c|c|cc|cc|cc}
    \toprule[0.5mm]
    \multirow{2}[2]{*}{Instance} & \multirow{2}[2]{*}{Scale} & \multicolumn{2}{c|}{LEHD-Greedy} & \multicolumn{2}{c|}{INViT-Greedy} & \multicolumn{2}{c}{L2R-Greedy} \\
          &       & Gap   & Time  & Gap   & Time  & Gap   & Time \\
    \midrule
    
    E10k.0 & 10,000 & 24.628\% & 12m   & 6.636\% & 3.8m  & \cellcolor[HTML]{D0CECE}\textbf{4.562\%} & \cellcolor[HTML]{D0CECE}\textbf{43s} \\
    E10k.1 & 10,000 & 26.506\% & 12m   & 7.080\% & 3.8m  & \cellcolor[HTML]{D0CECE}\textbf{4.591\%} & \cellcolor[HTML]{D0CECE}\textbf{44s} \\
    E10k.2 & 10,000 & 24.743\% & 12m   & 7.382\% & 3.8m  & \cellcolor[HTML]{D0CECE}\textbf{4.766\%} & \cellcolor[HTML]{D0CECE}\textbf{45s} \\
    E31k.0 & 31,623 & \multicolumn{2}{c|}{OOM} & 6.970\% & 29.2m & \cellcolor[HTML]{D0CECE}\textbf{4.822\%} & \cellcolor[HTML]{D0CECE}\textbf{2.4m} \\
    E31k.1 & 31,623 & \multicolumn{2}{c|}{OOM} & 7.222\% & 29.1m & \cellcolor[HTML]{D0CECE}\textbf{4.735\%} & \cellcolor[HTML]{D0CECE}\textbf{2.4m} \\
    E100k.0 & 100,000 & \multicolumn{2}{c|}{OOM} & \multicolumn{2}{c|}{N/A} & \cellcolor[HTML]{D0CECE}\textbf{4.711\%} & \cellcolor[HTML]{D0CECE}\textbf{8.2m} \\
    E100k.1 & 100,000 & \multicolumn{2}{c|}{OOM} & \multicolumn{2}{c|}{N/A} & \cellcolor[HTML]{D0CECE}\textbf{4.976\%} & \cellcolor[HTML]{D0CECE}\textbf{8.5m} \\
    E316k.0 & 316,228 & \multicolumn{2}{c|}{OOM} & \multicolumn{2}{c|}{OOM} & \cellcolor[HTML]{D0CECE}\textbf{4.820\%} & \cellcolor[HTML]{D0CECE}\textbf{27.1m} \\
    E1M.0 & 1,000,000 & \multicolumn{2}{c|}{OOM} & \multicolumn{2}{c|}{OOM} & \cellcolor[HTML]{D0CECE}\textbf{5.021\%} & \cellcolor[HTML]{D0CECE}\textbf{2.0h} \\
    E3M.0 & 3,162,278 & \multicolumn{2}{c|}{OOM} & \multicolumn{2}{c|}{OOM} & \cellcolor[HTML]{D0CECE}\textbf{5.049\%} & \cellcolor[HTML]{D0CECE}\textbf{12h} \\
    E10M.0 & 10,000,000 & \multicolumn{2}{c|}{OOM} & \multicolumn{2}{c|}{OOM} & \cellcolor[HTML]{D0CECE}\textbf{5.048\%} & \cellcolor[HTML]{D0CECE}\textbf{96h} \\
    \bottomrule[0.5mm]
    \end{tabular}%
    }
  \label{tab:dimacs}%
  \vspace{-15pt}
\end{table}%

\paragraph{Results on Benchmark Datasets}  
We further assess generalization on widely used benchmark datasets: CVRPLib Set-XXL~\citep{arnold2019cvrplib_xxl} and large-scale TSPLib~\citep{reinelt1991tsplib}. As shown in \cref{table:TSPLIB_CVRPLIB}, L2R remains the best-performing method across instances of varying scale, underscoring its practical applicability in real-world scenarios.  
\paragraph{Solving Very Large-Scale Instances}
\label{append:dimacs}
We further evaluate L2R on challenging large-scale TSP instances from the DIMACS challenge. As shown in \cref{tab:dimacs}, L2R can achieve strong performance on TSP instances with up to \textbf{10 million} nodes, which significantly outperforms other NCO methods. These results demonstrate that L2R, a purely learning-based constructive neural solver, consistently produces high-quality solutions for unprecedented problem instances with up to 10 million nodes. Notably, this capability is achieved by generalizing from 100-node training instances with uniform distributions, which represents a significant advancement in NCO. To the best of our knowledge, L2R is the first neural solver to effectively solve VRP instances with 10 million nodes.

\begin{table*}[t]
\centering
\caption{Evaluation of SSR strategies across benchmark datasets. "\#m" denotes the number of instances per range.}
\resizebox{0.8\linewidth}{!}{
\begin{tabular}{l|c|c|c|c|c|c}
\toprule[0.5mm]
\multirow{2}{*}{Method} & \multicolumn{3}{c|}{TSPLib} & \multicolumn{3}{c}{CVRPLib-XXL} \\
\cmidrule(lr){2-4} \cmidrule(lr){5-7}
 & 1K $\leq N \leq$ 5K (\#23) & 5K \textless $N \leq$ 100K (\#10)& All (\#33)& 3K $\leq N \leq$ 7K (\#4)& 7K \textless $N \leq$ 30K (\#6)& All (\#10)\\
\midrule
D-SSR greedy & 9.65\% & 7.47\% & 8.98\% & 14.38\% & 12.19\% & 13.06\% \\
L2R greedy & \cellcolor[HTML]{D0CECE}\textbf{9.16\%} & \cellcolor[HTML]{D0CECE}\textbf{7.36\%} & \cellcolor[HTML]{D0CECE}\textbf{8.61\%} & \cellcolor[HTML]{D0CECE}\textbf{12.05\%} & \cellcolor[HTML]{D0CECE}\textbf{11.22\%} & \cellcolor[HTML]{D0CECE}\textbf{11.55\%} \\
\bottomrule[0.5mm]
\end{tabular}
}
\label{table:ablation_lib_l2r_dssr}
\vspace{-5pt}
\end{table*}

\begin{figure*}[htbp]
    \centering
    \begin{subfigure}{0.24\textwidth}
        \centering
        \includegraphics[width=\textwidth]{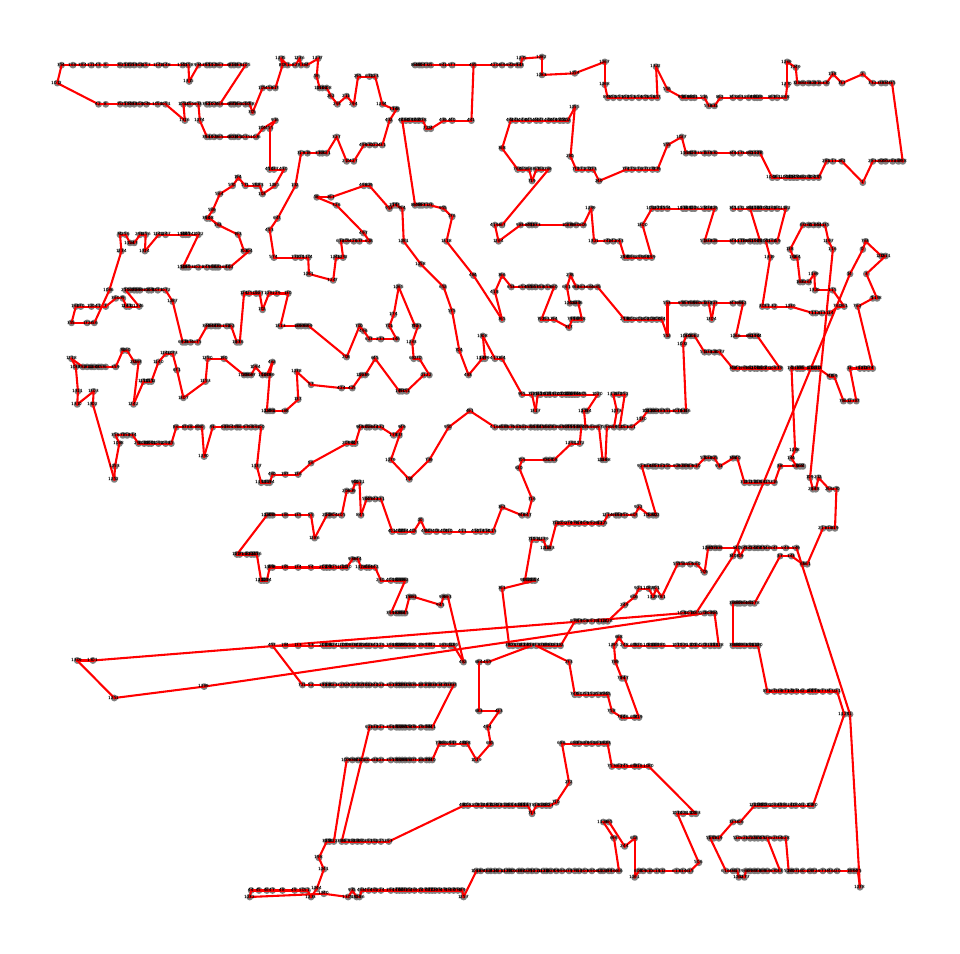}
        \caption{\centering D-SSR (rl1323)\\Gap: 16.44\%, Ratio: 93.88\%}
        \label{subfig:usa13509_dssr_sol}
    \end{subfigure}
     \begin{subfigure}{0.24\textwidth}
        \centering
        \includegraphics[width=\textwidth]{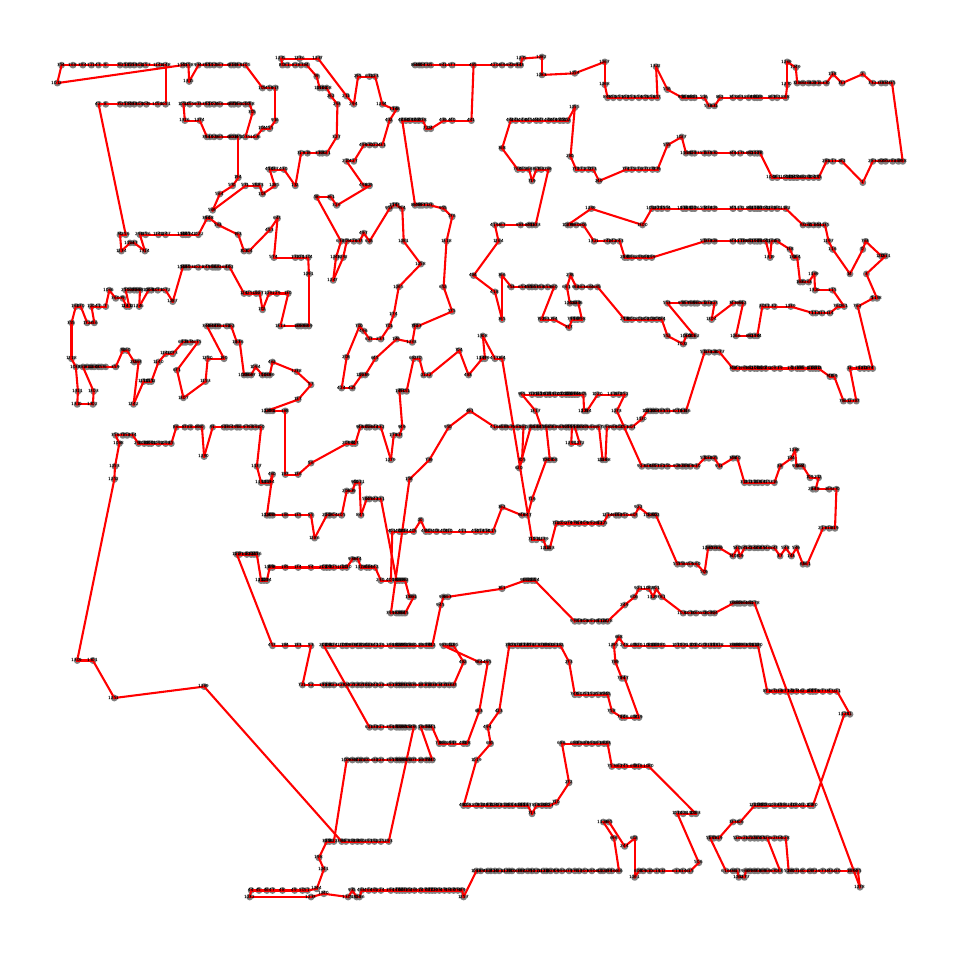}
        \caption{\centering L2R (rl1323) \\Gap: 9.34\%, Ratio: 94.63\%}
        \label{subfig:usa13509_l2r_sol}
    \end{subfigure}
    \begin{subfigure}{0.24\textwidth}
        \centering
        \includegraphics[width=\textwidth]{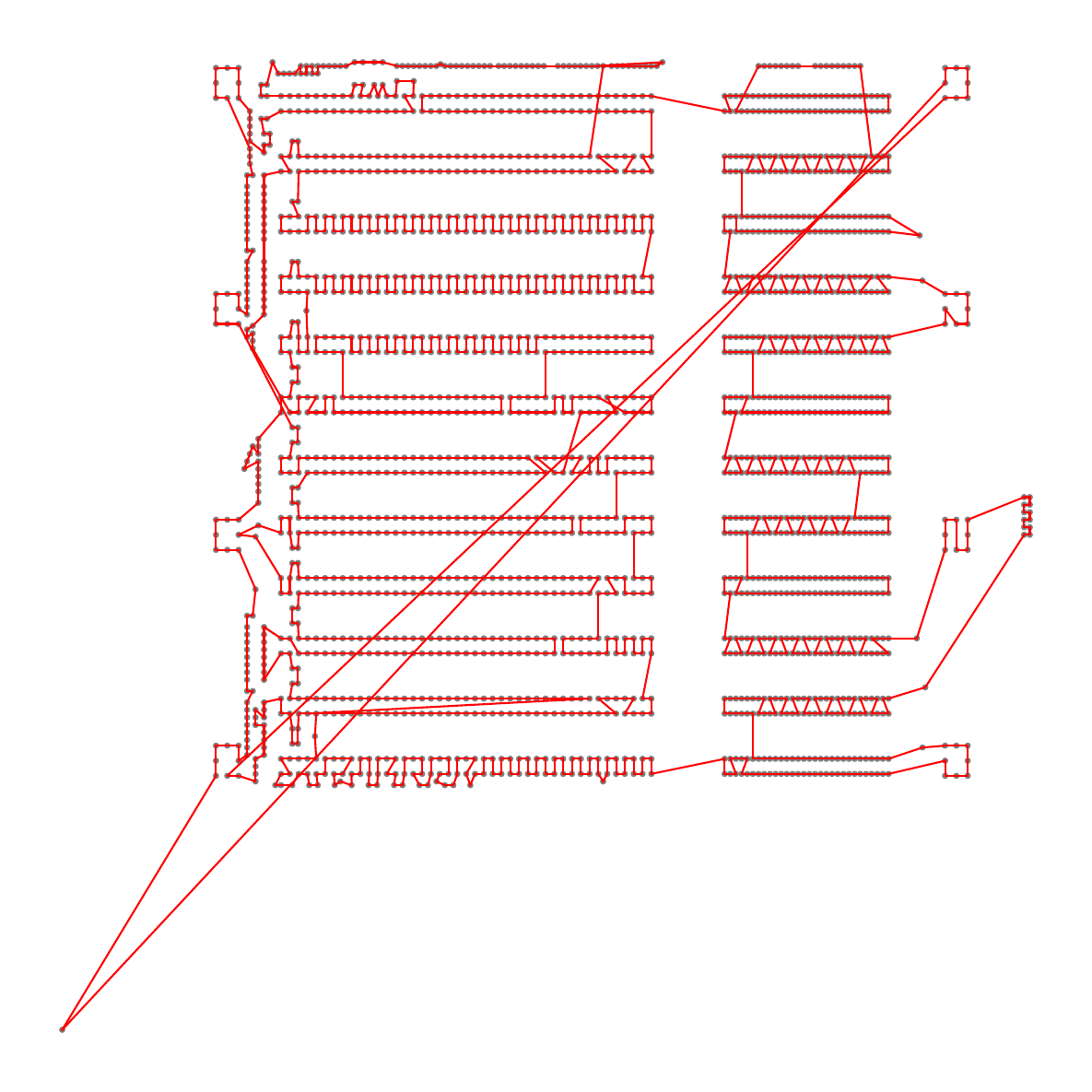}
        \caption{\centering D-SSR (d2103) \\ Gap: 20.50\%, Ratio: 94.53\%}
        \label{subfig:d2103_dssr_sol}
    \end{subfigure}
     \begin{subfigure}{0.24\textwidth}
        \centering
        \includegraphics[width=\textwidth]{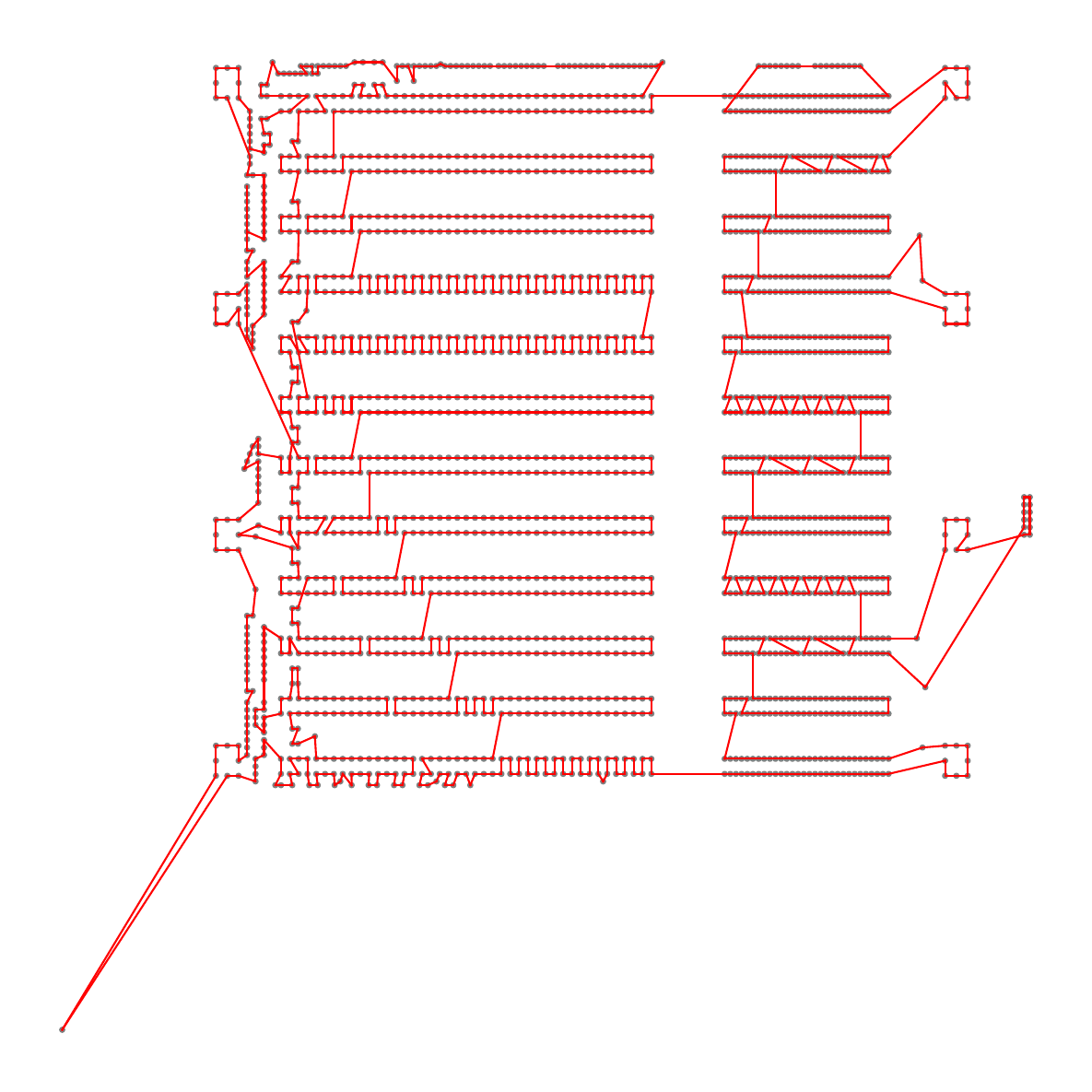}
        \caption{\centering L2R (d2103)\\Gap: 8.09\%, Ratio: 95.10\%}
        \label{subfig:d2103_l2r_sol}
    \end{subfigure}
\caption{Visualization of solutions for TSPLib instances of different scales generated by L2R and D-SSR, respectively. }
\label{fig:tsplib_analysis_visualizations}
\end{figure*}

\section{Further Analyses}
\subsection{L2R vs. D-SSR}
\label{sec:l2r_vs_dssr}
To validate the effectiveness of our proposed L2R, we evaluate it on large-scale TSPLib and CVRPLib benchmarks against D-SSR, a variant that replaces the learned reduction model with a distance-based $k$-NN rule while maintaining identical training settings. We systematically analyze the advantages of L2R, mainly including: (1) overall performance, (2) effects of different reduction strategies, and (3) instance-wise optimality gaps and ratios.

\paragraph{Overall Performance}
As shown in \cref{table:ablation_lib_l2r_dssr}, compared with traditional D-SSR, our proposed L2R achieves better generalization on large-scale TSPLib and CVRPLib instances. Notably, this advantage becomes more pronounced when solving complex CVRP instances.

\paragraph{Effects of Different Reduction Strategies}
To further support our argument, we visually compare the solutions generated by L2R and D-SSR. The optimality ratio is the proportion of nodes for which the optimal next-visit node is among the $k$ candidate nodes. As illustrated in \cref{fig:tsplib_analysis_visualizations}, while both D-SSR and L2R successfully select the majority of optimal next-visit nodes, D-SSR suffers a substantial degradation in overall solution quality compared to L2R. This degradation primarily stems from the fact that the distance-based SSR strategy eliminates non-local nodes (i.e., long-distance visits) that are essential for optimal routes, thereby forcing the local construction model into poor states and resulting in an anomalous next-visit node. This over-pruning effect accumulates systematically during the solution construction, ultimately compromising solution quality. In contrast, L2R's enhanced candidate selection capability enables it to identify more accurate candidate nodes, facilitating better node selections and ultimately achieving superior overall performance compared to D-SSR.

\begin{table}[t]
\centering
\caption{Optimality gap and optimality ratio per instance. "\#Accuracy" denotes the number of construction steps where $k$ candidates include the optimal next-visit node.}
\resizebox{0.5\textwidth}{!}{
\begin{tabular}{c| l | c | ccc |ccc }
\toprule[0.5mm]
&\multirow{2}{*}{Instance} & \multirow{2}{*}{Scale} &\multicolumn{3}{c|}{Gap $\downarrow$} & \multicolumn{3}{c}{\#Accuracy (Optimality Ratio) $\uparrow$} \\
&    &   &D-SSR  & L2R (Ours)&$\Delta$Gap $\downarrow$ &D-SSR &L2R (Ours)&$\Delta$\#Accuracy $\uparrow$ \\
\midrule
\multirow{10}{*}{\rotatebox{90}{TSPLib \hskip -1em}}
&pcb1173& 1173 & 5.17\%& \cellcolor[HTML]{D0CECE}\textbf{4.02\%} & $-$1.15\%&   1062 (90.54\%) & \cellcolor[HTML]{D0CECE}\textbf{1067 (90.96\%)} & +5\\
&rl1304& 1304 & 9.88\%& \cellcolor[HTML]{D0CECE}\textbf{8.14\%} & $-$1.74\%&  1225 (93.94\%) & \cellcolor[HTML]{D0CECE}\textbf{1234 (94.63\%)} & +9\\
&rl1323& 1323 & 16.44\%& \cellcolor[HTML]{D0CECE}\textbf{9.34\%} & $-$7.10\%& 1242 (93.88\%) & \cellcolor[HTML]{D0CECE}\textbf{1252 (94.63\%)} & +10\\ %mark
&u1817& 1817 & 17.60\%& \cellcolor[HTML]{D0CECE}\textbf{9.54\%} & $-$8.06\%&  1614 (88.83\%) & \cellcolor[HTML]{D0CECE}\textbf{1630 (89.71\%)} & +16\\%mark
&d2103& 2103 & 20.50\%& \cellcolor[HTML]{D0CECE}\textbf{8.09\%} & $-$12.41\%&  1988 (94.53\%) & \cellcolor[HTML]{D0CECE}\textbf{2000 (95.10\%)} & +12\\%mark
&pr2392& 2392 & 10.72\%& \cellcolor[HTML]{D0CECE}\textbf{8.08\%} & $-$2.64\%&  2105 (88.00\%) & \cellcolor[HTML]{D0CECE}\textbf{2129 (89.01\%)} & +24\\
&fnl4461& 4461 & 6.85\%& \cellcolor[HTML]{D0CECE}\textbf{4.22\%} & $-$2.63\%&  3871 (86.77\%) & \cellcolor[HTML]{D0CECE}\textbf{3896 (87.34\%)} & +25\\
&rl11849&11849  & 8.13\%& \cellcolor[HTML]{D0CECE}\textbf{6.96\%} & $-$1.17\%&  10851 (91.58\%) & \cellcolor[HTML]{D0CECE}\textbf{10864 (91.69\%)} & +13\\%mark
&usa13509& 13509 & 8.97\%& \cellcolor[HTML]{D0CECE}\textbf{6.99\%} & $-$1.98\%&  11886 (87.99\%) & \cellcolor[HTML]{D0CECE}\textbf{11947 (88.44\%)} & +61\\
&pla33810& 33810 & 7.57\%& \cellcolor[HTML]{D0CECE}\textbf{6.37\%} & $-$1.20\%&  30263 (89.51\%) & \cellcolor[HTML]{D0CECE}\textbf{30463 (90.10\%)} & +200\\
\midrule
\midrule
\multirow{7}{*}{\rotatebox{90}{CVRPLib \hskip -1em}}
&Leuven1 &3000  & 12.67\%& \cellcolor[HTML]{D0CECE}\textbf{11.49\%} & $-$1.18\%&  2284 (76.13\%) & \cellcolor[HTML]{D0CECE}\textbf{2294 (76.47\%)} & +10\\
&Leuven2 & 4000 & 15.71\%& \cellcolor[HTML]{D0CECE}\textbf{12.73\%} & $-$2.98\%&   3287 (82.18\%) & \cellcolor[HTML]{D0CECE}\textbf{3299 (82.48\%)} & +12\\%mark

&Antwerp1 &6000  & 13.12\%& \cellcolor[HTML]{D0CECE}\textbf{11.04\%} & $-$2.08\%&   4573 (76.22\%) & \cellcolor[HTML]{D0CECE}\textbf{4597 (76.62\%)} & +24\\%mark
&Antwerp2 & 7000 & 16.00\%& \cellcolor[HTML]{D0CECE}\textbf{12.95\%} & $-$3.05\%&   5611 (80.16\%) & \cellcolor[HTML]{D0CECE}\textbf{5638 (80.54\%)} & +27\\%mark
&Ghent1 &10000  & 11.68\%& \cellcolor[HTML]{D0CECE}\textbf{10.15\%} & $-$1.53\%&   7658 (76.58\%) & \cellcolor[HTML]{D0CECE}\textbf{7665 (76.65\%)} & +7\\%mark
&Ghent2 &11000  & 13.55\%& \cellcolor[HTML]{D0CECE}\textbf{11.29\%} & $-$2.26\%&   9024 (82.04\%) & \cellcolor[HTML]{D0CECE}\textbf{9045 (82.23\%)} & +21\\%mark
&Brussels2 &16000  & 13.83\%& \cellcolor[HTML]{D0CECE}\textbf{12.30\%} & $-$1.53\%&  13099 (81.87\%) & \cellcolor[HTML]{D0CECE}\textbf{13109 (81.93\%)} & +10\\%mark
\bottomrule[0.5mm]
\end{tabular}
}
\vspace{-15pt}
\label{table:analysis_tsplib_gap_ratio}
\end{table}

\paragraph{Optimality Gap \& Optimality Ratio}
We then measure the optimality gap and optimality ratio per instance. As demonstrated in \cref{table:analysis_tsplib_gap_ratio}, although L2R shows only marginal improvement in the optimality ratio, L2R achieves superior solution quality over D-SSR on real-world datasets. For D-SSR, the cumulative effect of over-pruning during solution construction leads to significant deviations from the optimal route in subsequent node selections, ultimately compromising the overall solution quality.

\subsection{Ablation Study}
\label{subsec:ablation_study}
We conduct detailed ablation studies to validate the effectiveness of L2R's core components, mainly including: (1) effects of adaptation bias in compatibility, (2) effects of different attention mechanisms, and (3) effects of static reduction. These results in Appendix \ref{append:ablation_study} confirm that the carefully designed components of our L2R framework are essential for robust and efficient generalization across three VRP variants while greatly reducing computational overhead without compromising solution quality.

\section{Conclusion}
\label{sec:conclusion}
In this work, we propose a novel RL-based L2R framework for large-scale VRPs. Unlike prior methods that rely on geometric distance as a rigid hard-pruning rule, L2R adaptively prioritizes nodes and prunes the search space by leveraging patterns learned from problem-specific features without compromising solution quality. Extensive experiments show that L2R generalizes robustly across diverse problem scales and data distributions on various VRPs. To the best of our knowledge, L2R is the first neural solver to effectively scale to VRP instances with $10$ million nodes while maintaining promising solution quality, significantly advancing the frontier of NCO in terms of generalization and scalability.

\begin{acks}
This work was supported by the National Natural Science Foundation of China (Grant No. 62476118), the Guangdong Provincial Key Laboratory of Fully Actuated System Control Theory and Technology (Grant No. 2024B1212010002), the Natural Science Foundation of Guangdong Province (Grant No. 2024A1515011759), and the Center for Computational Science and Engineering at Southern University of Science and Technology. 
\end{acks}
% \clearpage
\bibliographystyle{ACM-Reference-Format}
\balance
\bibliography{references}

\appendix
\section{Related Work}
\label{append:related_work}
\paragraph{NCO without Search Space Reduction}
Most NCO models are trained on small-scale instances (e.g., 100 nodes) without SSR and achieve strong performance on similarly sized instances. However, their effectiveness clearly diminishes when applied to larger instances (e.g., more than 1,000 nodes)~\citep{nazari2018reinforcement,xiao2024reinforcement,xin2021multidecoder,hottung2020nlns,chen2019NeuRewriter,luo2026rethinking}. Some approaches incorporate additional search procedures~\citep{deudon2018learning,bello2016neural,hottung2021eas} to mitigate this limitation. While improving solution quality, they are still computationally expensive. Another line of research focuses on training models directly on larger-scale instances (e.g., up to 500 nodes) to enhance generalization~\citep{wang2024distance,zhou2023omni,zhou2024icam}. However, this approach incurs prohibitive computational costs due to the exponentially growing search space. Alternatively, some methods simplify large-scale VRPs by decomposing them into smaller subproblems~\citep{kim2021_lcp,li2021_L2D,hou2022tam,pan2023htsp,zheng2024udc}. Although effective, the reliance on expert-designed policies limits their practical applications.

\paragraph{NCO with Static Search Space Reduction}
To address the scalability challenges, static SSR has been proposed as a computationally efficient approach. These methods perform one-time pruning at the beginning of the optimization process. For example, \citet{sun2021MLPR} develop a static problem reduction technique to eliminate unpromising edges in large-scale TSP instances. Recently, heatmap-based approaches have gained popularity for solving large-scale TSPs, in which models are trained to predict the probability that each edge belongs to the optimal solution. To handle large-scale instances, these methods often incorporate graph sparsification~\citep{qiu2022dimes,sun2023difusco,li2024T2T,li2024fastT2T} or pruning strategies~\citep{xing2020graph_mcts,fu2021attgcn_mcts,min2024utsp} to reduce the search space. While static SSR is computationally efficient, it typically requires well-designed search procedures (e.g., Monte Carlo Tree Search for TSP) to achieve high-quality solutions, which might be more important for the optimization process~\citep{xia2024position_mcts}. 
\paragraph{NCO with Dynamic Search Space Reduction}
Unlike static SSR, dynamic SSR adaptively prunes the search space to a small set of candidate nodes at each step, typically based on the distance to the last visited node. The final node selection can be guided by either the original policy augmented with auxiliary distance information~\citep{wang2024distance} or a well-designed local policy~\citep{gao2023elg}. Additionally, recent works~\citep{drakulic2023bq,fang2024invit,chen2026ttpl} directly select the next node from the candidate set using NCO models. While dynamic SSR can accelerate inference, its solely distance-based node selection struggles on instances featuring non-uniform distributions, or when optimal solutions rely heavily on non-spatial constraints.

\section{Local Solution Construction Model}
\label{append:local_construction_model}
The proposed model consists of four components detailed below: 1) coordinate normalization, 2) embedding layer, 3) attention layer, and 4) compatibility calculation.

\subsection{Coordinate Normalization}
\label{append:coor_norm}
Following \citep{fu2021attgcn_mcts,fang2024invit}, the coordinate transformation is formulated as
\begin{equation}
\begin{aligned}
    \Delta_{\max} &= \max\left(\max_{i \in \mathcal{N}_t} x_i - \min_{i \in \mathcal{N}_t} x_i, \ \max_{i \in \mathcal{N}_t} y_i - \min_{i \in \mathcal{N}_t} y_i\right), \\
    x_i^{\text{new}} &= \frac{x_i - \min_{j \in \mathcal{N}_t} x_j}{\Delta_{\max}}, \quad y_i^{\text{new}} = \frac{y_i - \min_{j \in \mathcal{N}_t} y_j}{\Delta_{\max}}, \quad \forall i \in \{\pi_1, \pi_{t-1}\} \cup \mathcal{N}_t.
\end{aligned}
\label{eq:coor_norm_direct}
\end{equation}
To ensure that the first visited node (or depot) $\pi_{1}$ remains within the boundary (i.e., $0 \leq x_{i}^{new} \leq 1$), we apply $x_{i}^{new} = \max(0, \min(x_{i}^{new}, 1))$, and the same operation is applied to $y_{i}^{new}$. Subsequently, the sub-graph $G_{\text{sub}}^{'}$ is transformed into a new graph $G_{\text{sub}}^{''}$. 

\subsection{Embedding Layer}
\label{append:local_embedding_layer}
\paragraph{TSP}  
Given the converted sub-graph $G_{\text{sub}}^{''}$, the embedding layer first transforms the coordinates into initial embeddings using a shared linear layer with learnable parameters $[W^{(0)} \in \mathbb{R}^{d_x \times d}; b^{(0)} \in \mathbb{R}^{d}]$. The embeddings of the $k$ candidate nodes $\mathcal{N}_{t}$ are denoted by $H_{\mathcal{N}}^{(0)} = \{ \mathbf{h}_i^{(0)} \mid i \in \mathcal{N}_{t} \} \in \mathbb{R}^{k \times d}$. Here, the first node $\pi_{1}$ and the last node $\pi_{t-1}$ are used to represent the current partial solution. Therefore, their initial embeddings require special treatment~\citep{drakulic2023bq,luo2023lehd}. Specifically, additional learnable matrices $W_{1} \in \mathbb{R}^{d \times d}$ and $W_{2} \in \mathbb{R}^{d \times d}$ are applied to $\mathbf{h}_{\pi_{1}}^{(0)}$ and $\mathbf{h}_{\pi_{t-1}}^{(0)}$, respectively. Accordingly, we define the initial graph node embeddings $\widetilde{H}^{(0)} =[W_{1}\mathbf{h}_{\pi_1}^{(0)}, \ W_{2}\mathbf{h}_{\pi_{t-1}}^{(0)}, \ H_{\mathcal{N}_{t}}^{(0)}]\in \mathbb{R}^{(2 + |\mathcal{N}_{t}|) \times d}$. Next, $\widetilde{H}^{(0)}$ is passed through the $M$ attention layers sequentially.

\paragraph{CVRP \& CVRPTW}  
For CVRP and CVRPTW, we adopt an attribute-aware layer to obtain the initial embeddings. The node-specific attributes $\bm{\omega}_{i}$ and a solution state vector $\bm{\rho}_{i}$ are defined as  
\begin{equation}
\bm{\omega}_{i} =
\begin{cases}
[\delta_i / Q_{\text{remain}}], \\
[\delta_i / Q_{\text{remain}}, e_{i}, l_{i}],
\end{cases}
\bm{\rho}_{i} =
\begin{cases}
[Q_{\text{remain}}], & \mathrm{CVRP},\\
[Q_{\text{remain}}, T_{\text{current}}], & \mathrm{CVRPTW},
\end{cases}
\label{eq:local_routing_attributes}
\end{equation}
where $\delta_i$, $e_i$, and $l_i$ are the node demand, start time, and end time of node $i$, respectively. $Q_{\text{remain}}$ and $T_{\text{current}}$ represent the remaining load and current time, respectively. We then project the coordinates and problem-specific attributes through separate linear layers:
\begin{equation}
\begin{aligned}
     \mathbf{h}_{i,\text{coor}}^{(0)} &= W^{(0)} [x_{i}, y_{i}] + \mathbf{b}^{(0)} \quad \forall i \in \{\pi_1, \pi_{t-1}\} \cup \mathcal{N}_{t}, \\
      \mathbf{h}_{i}^{(0)} &= \mathbf{h}_{i,\text{coor}}^{(0)} + W_{\text{attr}}\bm{\omega}_{i} \quad \forall i \in \mathcal{N}_{t}, \\
       \mathbf{h}_{\pi_{1}}^{(0)} &= W_{1} \mathbf{h}_{\pi_{1},\text{coor}}^{(0)} + W_{\text{state}}\bm{\rho}_{i}, \\
       \mathbf{h}_{\pi_{t-1}}^{(0)} &= W_{2} \mathbf{h}_{\pi_{t-1},\text{coor}}^{(0)} + W_{\text{state}}\bm{\rho}_{i},
\end{aligned}
\label{eq:local_capacitated_embedding}
\end{equation}
where $W_{\text{attr}}$ and $W_{\text{state}}$ are learnable matrices, with their input dimensions adapting to the feature size of the corresponding problem.

\subsection{Attention Layer}
\label{append:local_attention_layer}
Given the input $\widetilde{H}^{(\ell-1)} = [\mathbf{h}_{\pi_1}^{(\ell-1)}, \mathbf{h}_{\pi_{t-1}}^{(\ell-1)}, H_{\mathcal{N}_{t}}^{(\ell-1)}]$ to the $\ell$-th layer ($\ell = 1, \ldots, M$), the output for node $i$ is computed as:
\begin{equation}
\hat{\mathbf{h}}_i^{(\ell)} = \mathrm{LN}^{(\ell)}\left(\mathbf{h}_i^{(\ell-1)} + \mathrm{Attention}^{(\ell)}\left(\mathbf{h}_i^{(\ell-1)}, \widetilde{H}^{(\ell-1)}\right)\right),
\label{eq:attention_layer_aafm}
\end{equation}
\begin{equation}
    \mathbf{h}_i^{(\ell)} = \mathrm{LN}^{(\ell)}\left(\mathbf{\hat{h}}_i^{(\ell)} + \mathrm{FF}^{(\ell)}\left(\mathbf{\hat{h}}_i^{(\ell)}\right)\right),
\label{eq:FF_layer}
\end{equation}
where $\mathrm{LN}(\cdot)$ denotes layer normalization~\citep{ba2016layernorm}. $\mathrm{Attention}$ represents the adopted attention mechanism~\citep{zhou2024icam}, and $\mathrm{FF}(\cdot)$ indicates a fully connected neural network with ReLU activation. After $M$ attention layers, the final node embeddings $\widetilde{H}^{(M)} = [\mathbf{h}_{\pi_1}^{(M)}, \ \mathbf{h}_{\pi_{t-1}}^{(M)}, \ H_{\mathcal{N}_{t}}^{(M)}] \in \mathbb{R}^{(2 + |\mathcal{N}_{t}|) \times d}$ encapsulate the advanced feature representations of the first node, last node, and $k$ candidate nodes.

\subsection{Compatibility Calculation}
Similar to \cref{eq:u_calculated_reduction}, we compute the compatibilities $\mathbf{u}^{L}=\{u^{L}_{t,i}| i \in \mathcal{N}_{t}\}$ for selecting each node in $\mathcal{N}_{t}$. Specifically, we substitute $\hat{\mathbf{h}}_{C_D}^{t}$, $\mathbf{h}_{i}^{D}$, $a_{t-1,i}^{R}$, and the set $\mathcal{A}_t$ in \cref{eq:u_calculated_reduction} with $\hat{\mathbf{h}}_{(C)}^{t}$, $\mathbf{h}_{i}^{(M)}$, $a_{t-1,i}^{L}$, and $\mathcal{N}_t$, respectively. Here, $\hat{\mathbf{h}}_{(C)}^{t}=\mathbf{h}_{\pi_1}^{(M)}+\mathbf{h}_{\pi_{t-1}}^{(M)}$, and $a_{t-1,i}^{L} = -\alpha \cdot\log_2{|\mathcal{N}_{t}|}\cdot d_{t-1,i}$ is the adaptation bias between each node $i \in \mathcal{N}_{t}$ and $\pi_{t-1}$. Finally, the probabilities $\mathbf{p}=\{p_{i}|i \in \mathcal{N}_{t}\}$ are computed as $\mathbf{p} = \mathrm{softmax}(\mathbf{u}^{L})$.

\section{Ablation Study}
\label{append:ablation_study}
In this section, we conduct ablation studies to evaluate the effectiveness of each component. Please note that the results presented are obtained using greedy decoding.

\subsection{Effects of Adaptation Bias in Compatibility}
\label{append:ablation_compatibility}
To evaluate the effectiveness of two adaptation biases, we conduct an ablation study with different component configurations. The results in \cref{table:adaptation_bias_effect} demonstrate that accurately identifying candidate nodes within a large search space remains challenging when relying solely on a lightweight network architecture. Unlike existing distance-based reduction methods, our approach addresses the trade-off between learning difficulty and solution quality by incorporating a distance-assisted reduction model to enhance neighborhood selection performance. Both $a_{t-1,i}^{R}$ and $a_{t-1,i}^{L}$ contribute significantly to the promising generalization performance. 
\begin{table}[h!]
\centering
\caption{Ablation of the compatibility module. }
\vspace{-5pt}
\resizebox{0.8\columnwidth}{!}{
\begin{tabular}{ cc | c | c | c | c | c }
\toprule[0.5mm]
$a_{t-1,i}^{R}$ &$a_{t-1,i}^{L}$ & TSP1K & TSP5K & TSP10K & TSP50K & TSP100K  \\ 
\midrule
$\times$ &$\times$ & 53.91\% & 120.91\% & 157.46\% & 255.16\% & 305.95\%  \\ 
$\times$ &$\checkmark$ & 56.64\% & 129.57\% & 169.68\% & 279.63\% & 335.51\%  \\ 
$\checkmark$ &$\times$ & 5.22\% & 5.42\% & 5.66\% & 5.72\% & 5.68\%  \\ 
$\checkmark$ &$\checkmark$ &  \cellcolor[HTML]{D0CECE}\textbf{4.49\%} &   \cellcolor[HTML]{D0CECE}\textbf{4.69\%}  &  \cellcolor[HTML]{D0CECE}\textbf{4.82\%}  &   \cellcolor[HTML]{D0CECE}\textbf{4.86\%} &  \cellcolor[HTML]{D0CECE}\textbf{4.79\%} \\
\bottomrule[0.5mm]
\end{tabular}
}
\label{table:adaptation_bias_effect}
\vspace{-15pt}
\end{table}
\subsection{Effects of Different Attention Mechanisms}
\label{append:ablation_diff_attention}
We train an L2R variant that uses vanilla MHA~\citep{vaswani2017attention}, denoted L2R-MHA. Its training setup is identical to that of the original L2R, except for the use of attention. As shown in \cref{table:compare_AFT_MHA}, L2R-MHA shows robust large-scale generalization, further validating the effectiveness of our L2R framework. Notably, the original L2R yields additional performance improvements with shorter total runtime.
\begin{table}[htbp]
\centering
\caption{Comparison of different attention mechanisms.}
\vspace{-5pt}
\resizebox{0.8\columnwidth}{!}{
\begin{tabular}{ll | cc | cc | cc }
\toprule[0.5mm]
\multicolumn{2}{l|}{ }& \multicolumn{2}{c|}{TSP1K } & \multicolumn{2}{c|}{TSP10K }& \multicolumn{2}{c}{TSP100K } \\
\multicolumn{2}{l|}{Method}  & Gap & Time & Gap & Time  & Gap & Time \\
\midrule
\multicolumn{2}{l|}{L2R-MHA} & 6.90\% &  6.7s& 7.54\% &  1.1m& 7.72\% &  31.7m\\
\multicolumn{2}{l|}{L2R} &  \cellcolor[HTML]{D0CECE}\textbf{4.49\%} & \cellcolor[HTML]{D0CECE}\textbf{6.5s} &   \cellcolor[HTML]{D0CECE}\textbf{4.82\%} & \cellcolor[HTML]{D0CECE}\textbf{1.1m}  &  \cellcolor[HTML]{D0CECE}\textbf{4.79\%} &  \cellcolor[HTML]{D0CECE}\textbf{29.4m} \\
\bottomrule[0.5mm]
\end{tabular}
\vspace{-15pt}
}
\label{table:compare_AFT_MHA}
\end{table}

\subsection{Effects of Static Reduction}
\label{append:ablation_static_reduction}
Since strong baselines like BQ~\citep{drakulic2023bq} and INViT~\citep{fang2024invit} already incorporate distance-based dynamic reduction, we selected LEHD~\citep{luo2023lehd} as a baseline to verify the effect of static reduction. As shown in \cref{tab:ablation_static_reduction}, applying the same static reduction to LEHD yields a significant time reduction while maintaining solution quality. On the other hand, when removing SR from our L2R, performance remains unchanged, but the solving time increases, especially for TSP100K instances. This ablation study confirms that the static reduction serves as an effective, model-agnostic preprocessing step that improves efficiency without sacrificing solution quality. It also demonstrates that the superior performance of our L2R framework stems from its core architecture rather than just the preprocessing step. 

\begin{table}[htbp]
  \centering
  \caption{Comparison with and without static reduction for LEHD and our L2R. Note that "SR" is the static reduction technique we employed ($\gamma = 10\%$). }
  \vspace{-5pt}
  \resizebox{0.9\columnwidth}{!}{
    \begin{tabular}{c|cc|cc|cc}
    \toprule[0.5mm]
    \multirow{2}[1]{*}{Method} & \multicolumn{2}{c|}{TSP1K} & \multicolumn{2}{c|}{TSP10K} & \multicolumn{2}{c}{TSP100K} \\
          & Gap   & Time  & Gap   & Time  & Gap   & Time \\
    \midrule
    LEHD-Greedy w/o SR & 3.11\% & 98s   & 27.24\% & 3.1h  & \multicolumn{2}{c}{OOM} \\
    LEHD-Greedy w/ SR & 3.38\% & \cellcolor[HTML]{D0CECE}\textbf{73s} & 26.76\% & \cellcolor[HTML]{D0CECE}\textbf{2.3h} & \multicolumn{2}{c}{OOM} \\
    \midrule
    L2R-Greedy w/o SR & 4.49\% & 7s    & 4.82\% & 68s   & 4.79\% & 29.6m \\
    L2R-Greedy w/ SR & 4.49\% & \cellcolor[HTML]{D0CECE}\textbf{6s} & 4.82\% & \cellcolor[HTML]{D0CECE}\textbf{64s} & 4.79\% & \cellcolor[HTML]{D0CECE}\textbf{29.1m} \\
    \bottomrule[0.5mm]
    \end{tabular}%
    }
  \label{tab:ablation_static_reduction}%
  \vspace{-5pt}
\end{table}%
%%%%%%%%%%%%%%%%%%%%%%%%%%%%%%%%%%%%%%%%%%%%%%%%%%%%%%%%%%%%

\end{document}